\documentclass{article} 

\usepackage{url,hyperref,lineno,microtype}
\usepackage[onehalfspacing]{setspace}

\usepackage{soul}

\usepackage[USenglish]{babel}	
\usepackage{todonotes}
\usepackage{graphicx}
\usepackage{multirow}
\graphicspath{{figures/}}
    \DeclareGraphicsExtensions{.pdf,.eps,.png,.jpg}
\usepackage[caption=false,font=footnotesize]{subfig}
\usepackage{amsmath,amssymb}
\usepackage{bm}	
\usepackage[binary-units,per-mode=symbol]{siunitx}

\usepackage{booktabs}

\usepackage[noabbrev,capitalise]{cleveref} 
	\crefname{equation}{}{}	

\usepackage[nolist,smaller]{./texmf/tex/latex/common/eee-acronyms}
\usepackage{./texmf/tex/latex/common/acronym-patch,./texmf/tex/latex/common/acronym-abstract,./texmf/tex/latex/common/acronym-cap}
\usepackage{./texmf/tex/latex/common/eee-macros}
\usepackage{pdflscape}

\newcommand*{\Ninputs}{\ensuremath{L}}
\newcommand*{\Nclauses}{\ensuremath{U}}
\newcommand*{\Nclasses}{\ensuremath{M}}
\newcommand*{\Ntastates}{\ensuremath{2n}}
\newcommand*{\Nmidstate}{\ensuremath{n}}
\newcommand*{\Tainit}{\ensuremath{\phi_\textrm{Init}}}

\newcommand*{\True}{\ensuremath{1}}
\newcommand*{\False}{\ensuremath{0}}

\newcommand*{\pnoun}{\textsc}

\usepackage[nolist]{acronym}

  \begin{acronym}[DRAM]
    \acro{AI}{artificial intelligence}
    \acrodefindefinite{AI}{an}{an}
    \acro{ASIC}{application-specific integrated circuit}
    \acrodefindefinite{ASIC}{an}{an}
    \acro{BD}{bounded delay}
    \acro{BNN}{binarized neural network}
    \acro{CD}{completion detection}
    \acro{CNN}{convolutional neural network}
    \acro{CTM}{convolutional Tsetlin machine}
    \acro{CPOG}{conditional partial order graph}
    \acro{DI}{delay insensitive}
    \acro{DR}{dual-rail}
    \acro{DRAM}{dynamic RAM}
    \acro{DSP}{digital signal processing}
    \acro{DVFS}{dynamic voltage and frequency scaling}
    \acro{FD-SOI}{fully-depleted silicon-on-insulator}
    \acro{FPGA}{field-programmable gate array}
    \acro{RTL}{register transfer level}
    \acrodefindefinite{FPGA}{an}{a}
    \acro{FSM}{finite state machine}
    \acro{HDL}{hardware description language}
    \acrodefindefinite{HDL}{an}{a}
    \acro{INWE}{inverse-narrow-width effect}
    \acrodefindefinite{INWE}{an}{an}
    \acro{ISA}{instruction set architecture}
    \acro{IO}{input\slash output}
    \acrodefindefinite{ISA}{an}{an}
    \acro{IoT}{internet of things}
    \acrodefindefinite{IoT}{an}{an}
    \acro{LA}{learning automaton}
    \acrodefplural{LA}{learning automata}
    \acrodefindefinite{LA}{an}{a}
    \acro{LEC}{logical equivalence checking}
    \acrodefplural{LEC}{logical equivalence checks}
    \acro{LFSR}{linear feedback shift register}
    \acrodefindefinite{LFSR}{an}{a}
    \acro{LU}{learning unit}
    \acrodefindefinite{LU}{an}{a}
    \acro{MAC}{multiply-accumulate}
    \acro{MEP}{minimum energy point}
    \acrodefindefinite{MEP}{an}{a}
    \acro{ML}{machine learning}
    \acrodefindefinite{ML}{an}{a}
    \acro{MLP}{multi-layer perceptron}
    \acrodefindefinite{MLP}{an}{a}
    \acro{MPP}{maximum power point}
    \acrodefindefinite{MPP}{an}{a}
    \acro{MPPT}{maximum power point tracking}
    \acrodefindefinite{MPPT}{an}{a}
    \acro{MRAM}{}
    \acro{NCL}{null convention logic}
    \acrodefindefinite{NCL}{an}{a}
    \acro{NN}{neural network}
    \acrodefindefinite{NN}{an}{a}
    \acro{OCV}{on-chip variation}
    \acrodefindefinite{OCV}{an}{an}
    \acro{PCG}{permuted congruential generator}
    \acro{PRNG}{psuedorandom number generator}
    \acro{PVT}{process, variation and temperature}
    \acro{QDI}{quasi delay insensitive}
    \acro{RAM}{random-access memory}
    \acrodefplural{RAM}{random-access memories}
    \acro{RDF}{random dopant fluctuation}
    \acro{RTM}{recurrent Tsetlin machine}
    \acrodefindefinite{RTM}{an}{a}
    \acro{SCM}{standard cell memory}
    \acrodefindefinite{SCM}{an}{a}
    \acrodefplural{SCM}{standard cell memories}
    \acro{SI}{speed independent}
    \acrodefindefinite{SI}{an}{a}
    \acro{SR}{single-rail}
    \acrodefindefinite{SR}{an}{a}
    \acro{SRAM}{static RAM}
    \acrodefindefinite{SRAM}{an}{a}
    \acro{STG}{signal transition graph}
    \acrodefindefinite{STG}{an}{a}
    \acro{SVM}{support vector machine}
    \acrodefindefinite{SVM}{an}{a}
    \acro{TA}{Tsetlin automaton}
    \acrodefplural{TA}{Tsetlin automata}
    \acro{TAT}{Tsetlin automaton team}
    \acro{TM}{Tsetlin machine}
    \acro{ULV}{ultra-low voltage}
    \acrodefindefinite{ULV}{a}{an}
    \acro{VLSI}{very-large-scale integration}
    \acro{WSN}{wide sensor network}
    \acro{IoT}{Internet of Things}
    \acro{IC}{integrated circuit}
  \end{acronym}

\allowdisplaybreaks

\def\firstAuthorLast{Shafik {et~al.}} 

\def\Address{{$^\dagger$}Microsystems Group,
School of Engineering, Newcastle University, NE1 7RU, UK.\\
{$^\ddagger$}Centre for AI Research (CAIR), University of Agder, Kristiansand, Norway.\\
{$^\textbullet$}Valencia Polytechnic University, Spain.\\
{$^{**}$}The University of Edinburgh, UK.}
\def\corrAuthor{Rishad Shafik} \def\corrEmail{Rishad.Shafik@newcastle.ac.uk}

\begin{document}
\vspace*{0.35in}

\begin{flushleft}
{\Large

\textbf{\textit{Energy-frugal} and \textit{Interpretable} AI Hardware Design using Learning Automata} 
}
\newline
\\
Rishad Shafik\textsuperscript{1,*},
Tousif Rahman\textsuperscript{1},
Adrian Wheeldon\textsuperscript{2},
Ole-Christoffer Granmo\textsuperscript{3},
Jie Lei\textsuperscript{4},
Alex Yakovlev\textsuperscript{1}
\\
\bigskip
\bf{1} Microsystems Group, School of Engineering, Newcastle University, UK.
\\
\bf{2} The University of Edinburgh, UK.
\\
\bf{3} The Centre for AI Research (CAIR), University of Agder, Norway.
\\
\bf{4} Valencia Polytechnic University, Spain.
\\
\bigskip
* correseponding@author.mail

\end{flushleft}


\title{\textit{Energy-frugal} and \textit{Interpretable} AI Hardware Design using Learning Automata} 



\begin{abstract}

Energy efficiency is a crucial requirement for enabling powerful \ac{AI} applications at the microedge. Hardware acceleration with frugal architectural allocation is an effective method for reducing energy. Many emerging applications also require the systems design to incorporate interpretable decision models to establish responsibility and transparency. The design needs to provision for additional resources to provide reachable states in real-world data scenarios, defining conflicting design tradeoffs  between energy efficiency.  is challenging.

Recently a new machine learning algorithm, called the Tsetlin Machine, has been proposed. The algorithm is fundamentally based on the principles of finite-state automata and benefits from natural logic underpinning rather than arithmetic. In this paper, we investigate methods of energy-frugal \ac{AI} hardware design by suitably tuning the hyperparameters, while maintaining high learning efficacy. To demonstrate interpretability, we use reachability and game-theoretic analysis in two simulation environments: a SystemC model to study the bounded state transitions in the presence of hardware faults and Nash equilibrium between states to analyze the learning convergence. Our analyses provides the first insights into conflicting design tradeoffs involved in energy-efficient and interpretable decision models for this new \ac{AI} hardware architecture. We show that frugal resource allocation coupled with systematic prodigality between randomized reinforcements can provide decisive energy reduction while also achieving robust and interpretable learning.

\tiny
\end{abstract}

	\section{Introduction}

Minimizing energy consumption is a primary design objective in embedded \ac{AI} applications, such as image and voice recognition~\cite{shafik2018real,mathew2014energy,beninirtns17,shafik2009soft}. In many applications, hardware acceleration is preferred over software implementation as the former is significantly more energy efficient~\cite{garofalo2020pulp}. To reduce energy in hardware, architectural resource pruning is an effective method which aims to cut down the non-critical data computation and movement costs. Examples in traditional \acp{NN} include approximate arithmetic design~\cite{ansari2019improving, qiqieh2018significance}, network sparsification~\cite{lim2021spontaneous, conti2018xnor}, network compaction using hyperparameter search~\cite{bergstra2012random} and mixed signal design~\cite{mileiko2020neural, serb2018seamlessly}. However, architectural changes such as these affect the learning accuracy and make the decision process sensitive to parametric and data variations~\cite{wang2019deep, zhang2020survey}. 

Interpretability is another property of the \ac{AI} design that allows for explaining the method of learning (\ie{} training) and the decision models (\ie{} classification) from data in a humanly intelligible way~\cite{bhatt2020explainable}. It is an important design objectives to establish responsibility in autonomous applications, particularly for those that are safety-critical. Currently there is growing interest in interpretable \ac{AI} systems design implemented using \acp{NN}. However, this has remained non-trivial for their complex arithmetic underpinning, including variable gradient-descent based learning behavior during the training regime~\cite{gunning2017explainable,doran2017does,yampolskiy2019unexplainability}. This is further exacerbated by design approximation methods for energy efficiency~\cite{wang2019deep}.

\Acp{LA} constitutes a class of \ac{ML} with unique discrete reinforcement characteristics that can address the above design objectives~\cite{granmo2007learning}. Originally proposed by Mikhail Tsetlin, it uses the finite-state automaton as the basic learning unit~\cite{gel1962some}. Each automaton reinforces the current action using the past history, following the trajectory of a probability distribution. This probability distribution is updated based on the environmental responses the automaton obtains by performing a particular action. However, as the number of actions, and their probability distribution trajectories have a very large number of combinations, designing compact decision systems using \ac{LA} has been challenging~\cite{Oommen1988,Narendra74}. 

Recently, \ac{TM} has been proposed as a promising new \ac{ML} algorithm~\cite{Granmo2018} that simplifies the traditional \aclp{LA} by combining the state-bounded action updates with contemporary game theory~\cite{von2007theory}. Each automaton, defined as the finite automata with linear tactics or \ac{TA}, can independently ``play games'', \ie{} update its internal states and actions, using the newly refined reinforcement mechanisms (see~\cref{tab:feedback}). These have enabled the formulation of a learning problem through hierarchical and powerful propositional logic expressions~\cite{rahimi2019tsetlin,Maheshwari2023,lei2020arithmetic}. Exploiting these the first-ever \ac{TM} hardware architecture was proposed, which demonstrated significantly higher energy efficiency than state-of-the-art \acp{NN}~\cite{Wheeldon2020a}. A brief description of the Tsetlin Machine is provided in~\cref{sec:TMintro}.

The efficacy of hardware \aclp{TM} depends on a number of hyperparameters (see \cref{tab:hyperparams}). These are often inter-dependent and intertwined due to stochastic nature of propositional logic based learning within the reinforcement components of the \ac{TM} algorithm~\cite{rahimi2019tsetlin}. For understanding the relationships between them and thereby achieving the maximum efficacy, a systematic design space exploration is needed.

In a \ac{TM}, when the energy efficiency objective is coupled with learning efficacy, this can indeed be significantly more challenging. On the one hand, energy-frugality favors using the least amount of resources (\eg{} the minimum number of clauses and reinforcement events per learning epoch). On the other hand, accuracy requires significantly higher stochastic diversity between the reinforcement components (using higher number of clauses as well as concurrent learning events). To determine a balance between these conflicting requirements, designing a suitable prodigality is essential. Prodigality allows the system to navigate through the maximum number of bounded state transitions and tries not to miss its best scenarios. One key mechanism for achieving prodigality in \ac{TM} is to enable the system to perturb its the state transitions with an aim to performing the optimal number of reinforcement steps.

In this paper, we investigate the method of leveraging the \ac{LA} hyperparameters to resolve this natural conflict in the best possible way. For that we systematically study reachability under controlled redundancy and randomization. Further, we analyze the state reachability and convergence in a game-theoretic setting (with Nash equilibria). Our overall aim is to demonstrate the methods of energy-frugal and explainable \ac{ML} designs that are pivotal for the growth of this new \ac{AI} algorithm.


The paper is organized as follows. Section~\ref{sec:tsetlin} introduces \ac{LA} hyperparameters in the context of hardware architecture. Section~\ref{sec:design} studies their impact on the conflicting tradeoffs between energy, accuracy and performance. Sections~\ref{sec:reachability} and~\ref{sec:game_theoretic} provide a state transition based reachability and convergence analysis with and without faults present. Finally, Section~\ref{sec:conclusions} summarizes the analysis and discusses future work.

\section{The Tsetlin Machine} \label{sec:TMintro}
\begin{figure*}[htb]
    \centering
    \includegraphics[width=0.8\textwidth]{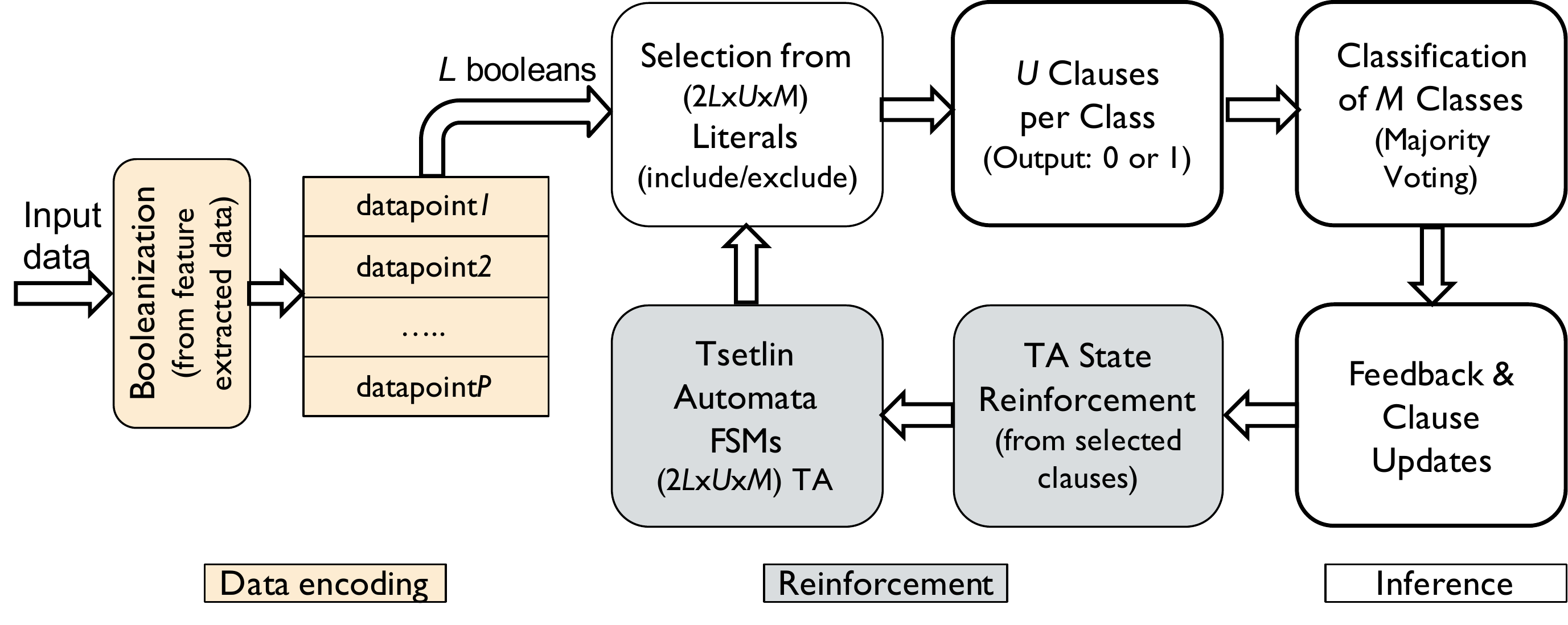}
    \caption{A schematic diagram of \ac{TM}, showing 3 different structural components: reinforcement components primarily used in the learning\slash training process, inference components which are crucial for classification and data encoding that allows parallel inputs both during training and inference.}
    \label{fig:tsetlin}
\end{figure*}
Figure~\ref{fig:tsetlin} depicts a schematic diagram of the \ac{TM} hardware architecture consisting of 3 structural components~\cite{Granmo2018, Wheeldon2020a}: data encoding, reinforcement and inference. These are briefly described below.

\noindent \textbf{Data encoding}: \acp{TM} encode the input data as a set of Boolean digits with equal significance, which we term as \textit{Booleanization}. It is different from Binarization in \acp{NN}, where data are encoded in binary numbers with positional significance of digits. Details of the method of Booleanization are outside the scope of this paper; interested readers can refer to~\cite{Wheeldon2020a}. The encoded Boolean digits and their complements define a set of input literals for the \ac{TM}. Each clause, representing a propositional logic unit, consists of all input literals and their corresponding \acp{TA} --- with a selected number of literals included in the conjunctive logic expression. The number of clauses in each class is an important design parameter that depends on complexity of the \ac{ML} problem and is determined at design time.

\noindent \textbf{Reinforcement}: The combination of input literals are then included or excluded in the clause output definition depending on the internal states of the reinforcement components, \acp{TA}. These are the finite automata with linear tactics, implemented as \acp{FSM} in the hardware. Each automaton constitutes a set of states that define the discrete action space. During training rewards are used to reinforce the states towards an action and penalties are used to transition the states for weakening automaton confidence in performing an action. The action updates take place in discrete space, rather than in gradient-descent steps,  which is a major differentiator when compared to traditional neural networks. This feature can be exploited for discernible and explainable \ac{AI} hardware design. This requires understanding reachability of \acp{TA} states and clause outputs in relation to the Boolean literals during the training and inference exercises. 

\noindent \textbf{Inference}: After training, an ensemble of \ac{TA} states (\ie{} their actions) define the selection of the literals as well as the output of a clause. To implement a propositional structure, the clauses are divided in two groups: positive clauses and negative clauses. Using a majority voting mechanism, the group of clauses with the most number of logic 1 outputs is able to infer a class definition. 

A detailed introduction to \ac{TM} hardware architecture and the original algorithm can be found here~\cite{Granmo2018, Wheeldon2020a}.
	\section{Tsetlin Machine Hyperparameters} \label{sec:tsetlin}
In the following we briefly introduce the architectural and learning hyperparameters that affect learning efficacy, energy and reachability. \\

\subsection{Architectural Hyperparameters}
\Ac{TM} is intrinsically hierarchical. Both reinforcement and inference components have the same number of stochastically reinforced clauses for each output class. The output of these clauses is defined by a team of \ac{TA}. The number of \ac{TA} is determined by the number of booleanized digits and their complements (which can be application specific~\cite{rahman2022data}). Each automaton consists of $2n$ states with $n$ being the decision boundary between actions: \textit{include} and \textit{exclude}. The choice of $2n$ influences the register sizing within each automaton \ac{FSM} as well as the number of reachable state transitions for convergence (see Section~\ref{sec:reachability}). 

Typically, for an application the numbers of booleanized digits ($\Ninputs{}$) and output classes ($\Nclasses{}$) are pre-defined and fixed. The designer then suitably allocates $\Nclauses{}$ clauses per output class. Thus, a total of $2\Ninputs{}\times \Nclauses{} \times \Nclasses{}$ \ac{TA} are needed. Choosing the number of clauses is non-trivial as there are conflicting design tradeoffs between learning accuracy and energy efficiency (see Section~\ref{sec:design}). Higher number of clauses offers more stochastic diversity in the propositional logic and as such favors better learning accuracy, while affecting the system energy consumption. Conversely, lower number of clauses reduces the energy consumption at the cost of less stochastic diversity and thereby degraded learning efficacy (see Section~\ref{sec:design}).
\begin{table*}[htbp]
\small{
\centering
\caption{Reinforcement feedback in a \ac{TA} deciding on \emph{include} or \emph{exclude} of a given literal $l_k$ in the clause $C^{i+}_j$. NA refers to \textit{no action}.}
\begin{tabular}{c|ccccc|cccc}
\multicolumn{2}{r|}{{\bf Feedback Type}} & \multicolumn{4}{c}{{\bf Type I}}&\multicolumn{4}{c}{\bf{Type II}}\\\hline
\multicolumn{2}{r|}{{\it Truth Value of Clause} $C_j^{i+}$ }&\multicolumn{2}{c}{\True}&\multicolumn{2}{c|}{\False} & \multicolumn{2}{c}{\True}&\multicolumn{2}{c}{\False}\\  
\multicolumn{2}{r|}{{\it Truth Value of Literal} $l_k$}&{\True}&{\False}&{\True}&{\False}&{\True}&{\False}&{\True}&{\False}\\
 \hline
 \hline
    \multirow{3}{*}{\bf Include Literal ($l_k \in L_j^{i+}$)}&\multicolumn{1}{c|}{$P(\mathrm{Reward})$}&$\frac{s-1}{s}$&NA&$0$&$0$&$0$&$\mathrm{NA}$&$0$&$0$\\
    &\multicolumn{1}{c|}{$P(\mathrm{Inaction})$}&$\frac{1}{s}$&NA&$\frac{s-1}{s}$&$\frac{s-1}{s}$&$1.0$&$\mathrm{NA}$&$1.0$&$1.0$\\
  &\multicolumn{1}{c|}{$P(\mathrm{Penalty})$}&$0$&NA&$\frac{1}{s} $&$\frac{1}{s}$&$0$&$\mathrm{NA}$&$0$&$0$\\
  \hline
  \multirow{3}{*}{\bf Exclude Literal ($l_k \notin L_j^{i+}$)}&\multicolumn{1}{c|}{$P(\mathrm{Reward})$}&$0$&$\frac{1}{s}$&$\frac{1}{s}$ &$\frac{1}{s}$&$0$&$0$&$0$&$0$\\
  &\multicolumn{1}{c|}{$P(\mathrm{Inaction})$}&$\frac{1}{s}$&$\frac{s-1}{s}$&$\frac{s-1}{s}$ &$\frac{s-1}{s}$&$1.0$&$0$&$1.0$&$1.0$\\
  &\multicolumn{1}{c|}{$P(\mathrm{Penalty})$}&$\frac{s-1}{s}$&$0$&$0$&$0$&$0$&$1.0$&$0$&$0$\\
  \hline
\end{tabular}
\label{tab:feedback}
}
\end{table*}

\subsection{Learning Hyperparameters}
With the given architectural parameters, the actual learning process in \ac{TM} involves a game theory inspired randomized state transitions between the automata~\cite{Narendra-book}. In each round of the game, a selection of these automata independently decide the next state transitions and actions within their respective clauses. This selection process is governed by the following feedback steps in the iterative learning process, as follows: 
\begin{enumerate}
    \item The current automata actions (\textit{include} or \textit{exclude}) form a propositional logic expression between the literals, which defines the clause output using the training datapoint (which is a set of booleanized literals used in training or inference).
    \item For a given datapoint, the \ac{TM} architecture then generates two groups of clause outputs with equal number of clauses: positive polarity clauses ($C_j^{i+}$) and negative polarity clauses ($C_j^{i-}$). By deducing the sum of all ($C_j^{i-}$) from the sum of all ($C_j^{i+}$) the learning classification is produced~\cite{Granmo2018}.
    \item If the output is false negative or true positive (\ie expected output is 0 or 1 but the current output is 1), then type I feedback is required for automaton corresponding to the literal $l_k$ within a $C_j^{i+}$ (\cref{tab:feedback}).
    \item If the output is false positive (\ie expected output is 1 but the current output is 0), then type II feedback is required for automaton corresponding to the literal $l_k$ within a $C_j^{i+}$ (\cref{tab:feedback}).
\end{enumerate}

The selection of the clauses that are reinforced in steps 3 and 4 above depends on a feedback threshold ($T$). A higher $T$ value forces a larger randomly selected team of clauses to participate in the reinforcement process providing with more stochastic diversity in the reinforcement steps or vice versa. Each automaton within the selected clauses also updates the state transitions based on the stochastic variable called learning sensitivity ($s$), which controls the level of agility in issuing a reward or penalty to each \ac{TA}. Random number generation is crucial for controlling these stochastic variables: $s$ and $T$. With suitably chosen hyperparameters, rewards and penalties allow for the state transitions, while no action is crucial for learning stability. 

\begin{table*}[ht]
\small{
  \centering
  \caption{%
    \Acl{TM} parameters, their symbols and system-wide impact.
  }\label{tab:hyperparams}
  \begin{tabular}{*3l}
    \textbf{Hyperparameter} & \textbf{Symbol} & \textbf{Impact}\\
    \toprule
    Number of booleanized inputs & \Ninputs{} & Application-specific and fixed \\
    Number of classes & \Nclasses{} & Application-specific and fixed\\
    Number of clauses per class & \Nclauses{} & Influences accuracy\slash energy\slash prodigality \\
    Number of automaton states & \Ntastates{} & Influences reachability and \ac{FSM} architecture\\
    Automaton decision boundary & \Nmidstate{} & Influences reachability and \ac{FSM} architecture \\
    Automaton initialization state & \Tainit{} & Influences reachability (typically \Tainit{}=$S_n$ or $S_{n+1}$) \\
    Feedback threshold & $T$ & Influences concurrent learning events of clauses\\
    Learning Sensitivity & $s$  & Influences the penalty\slash reward probabilities in automaton\\
    \bottomrule
  \end{tabular}
}
\end{table*}
\Cref{tab:hyperparams} summarizes the \acl{TM} hyperparameters with their associated symbols and impacts. In the next section their impact on energy-frugality and prodigality will be investigated in detail.

	\begin{figure*}[htbp]
    \centering
    \includegraphics[width=0.9\textwidth]{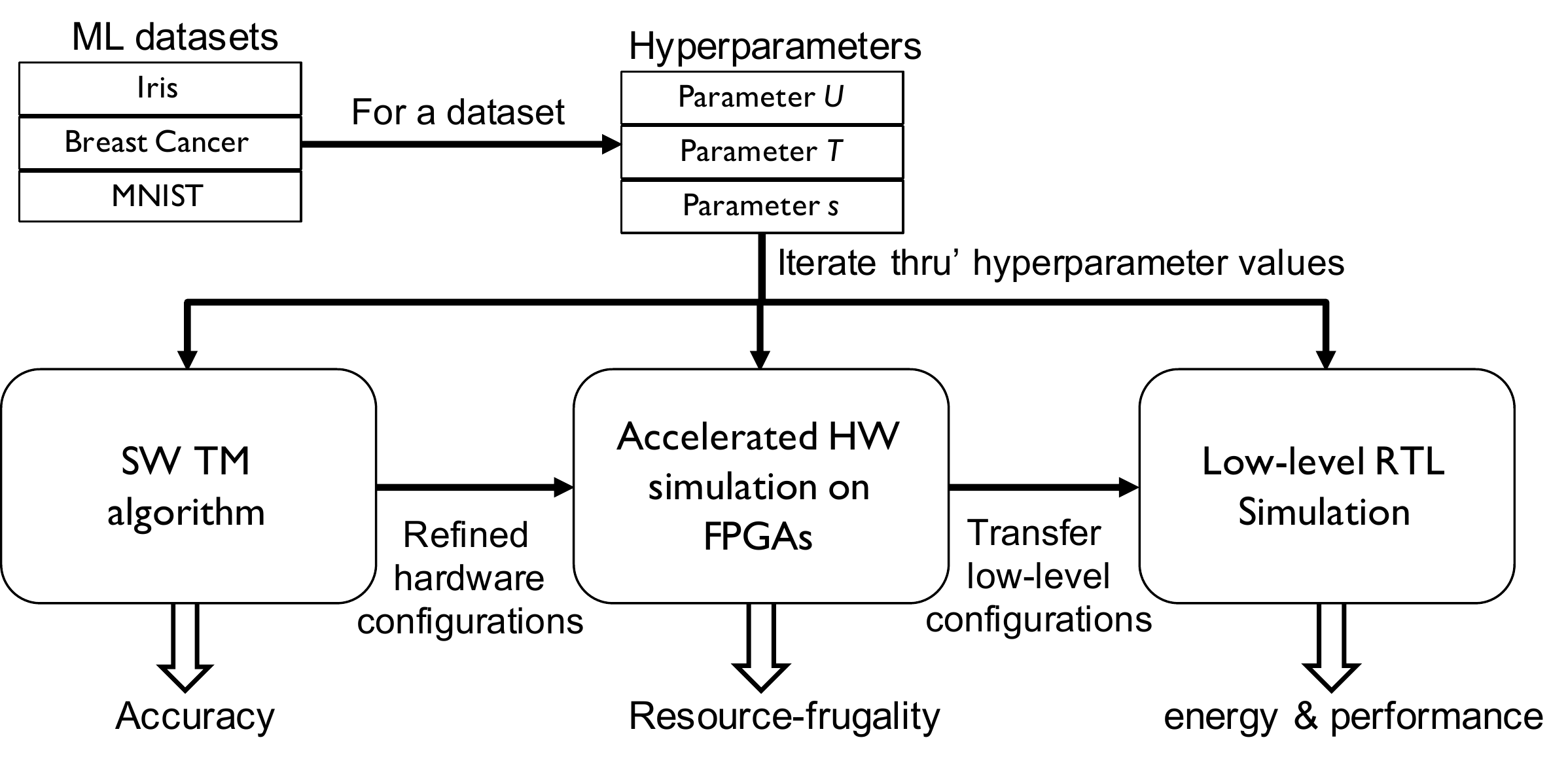}
    \caption{A hardware\slash software co-design framework for \ac{TM} design automation and exploration. The aim is to achieve energy-frugality, performance and accuracy in the hardware \ac{TM}.}
    \label{fig:design}
\end{figure*}

\section{Energy-Frugal Design} \label{sec:design}

\subsection{Design Exploration Framework} \label{sec:design:automation}

\Cref{fig:design} shows the hardware\slash software co-design framework used in our design exploration exercises. In this framework, software (SW) models are used for validating training and inference accuracy, typically for the larger \ac{ML} datasets. However, they have limited low-level hardware validation capability. \Acp{FPGA} prototype models are used to facilitate accelerated design explorations with refined hardware configurations. These are ideal for resource-frugality as well as hardware-level learning accuracy validations. However, \acp{FPGA} have limited flexibility and scalability of \ac{ML} datasets and as such scaled down datasets are used. By transferring the configurations onto \ac{RTL} models, low-level hardware prototype is designed, which are validated for high-fidelity figures of energy-frugality, accuracy and performance. Since low-level hardware simulations are computationally expensive, FPGA prototype models are iterated with different parametric values for faster design exploration. 
\begin{figure*}[htbp]
    \centering 
\subfloat[Iris]{\includegraphics[width=0.32\textwidth]{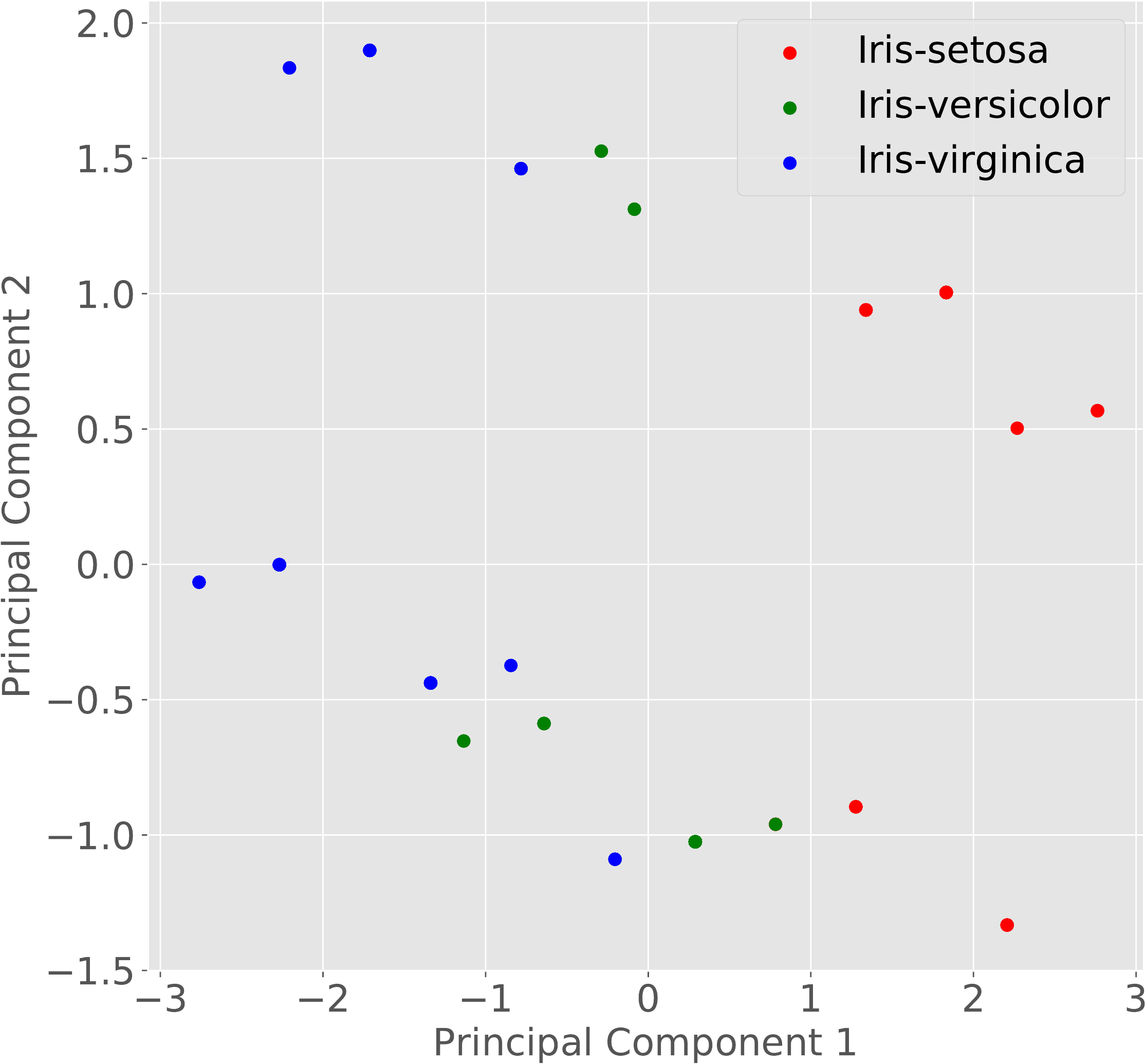}\label{fig:pca_iris}}
\subfloat[Breast Cancer]{\includegraphics[width=0.32\textwidth]{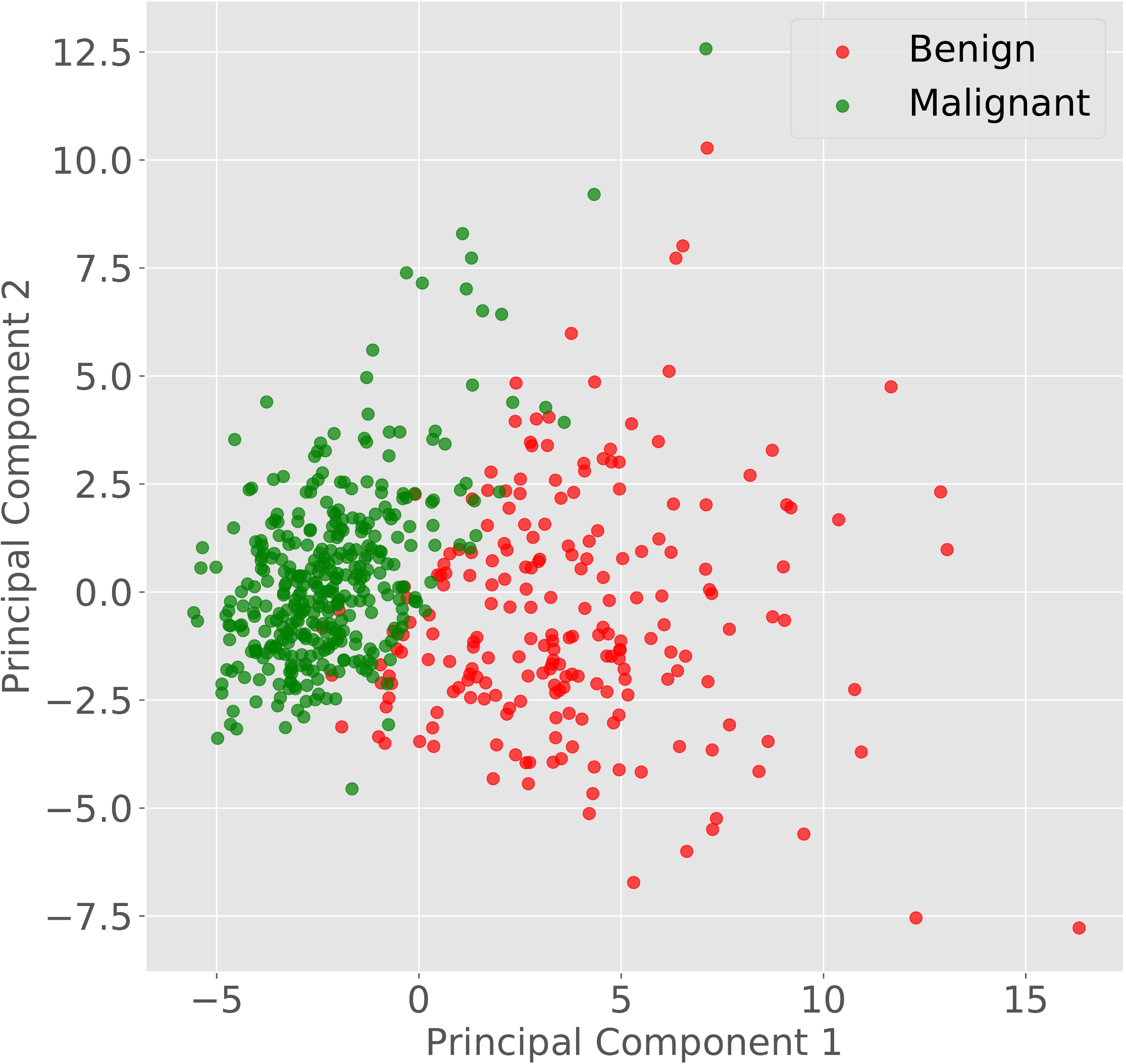}\label{fig:pca_bc}}
\subfloat[MNIST]{\includegraphics[width=0.32\textwidth]{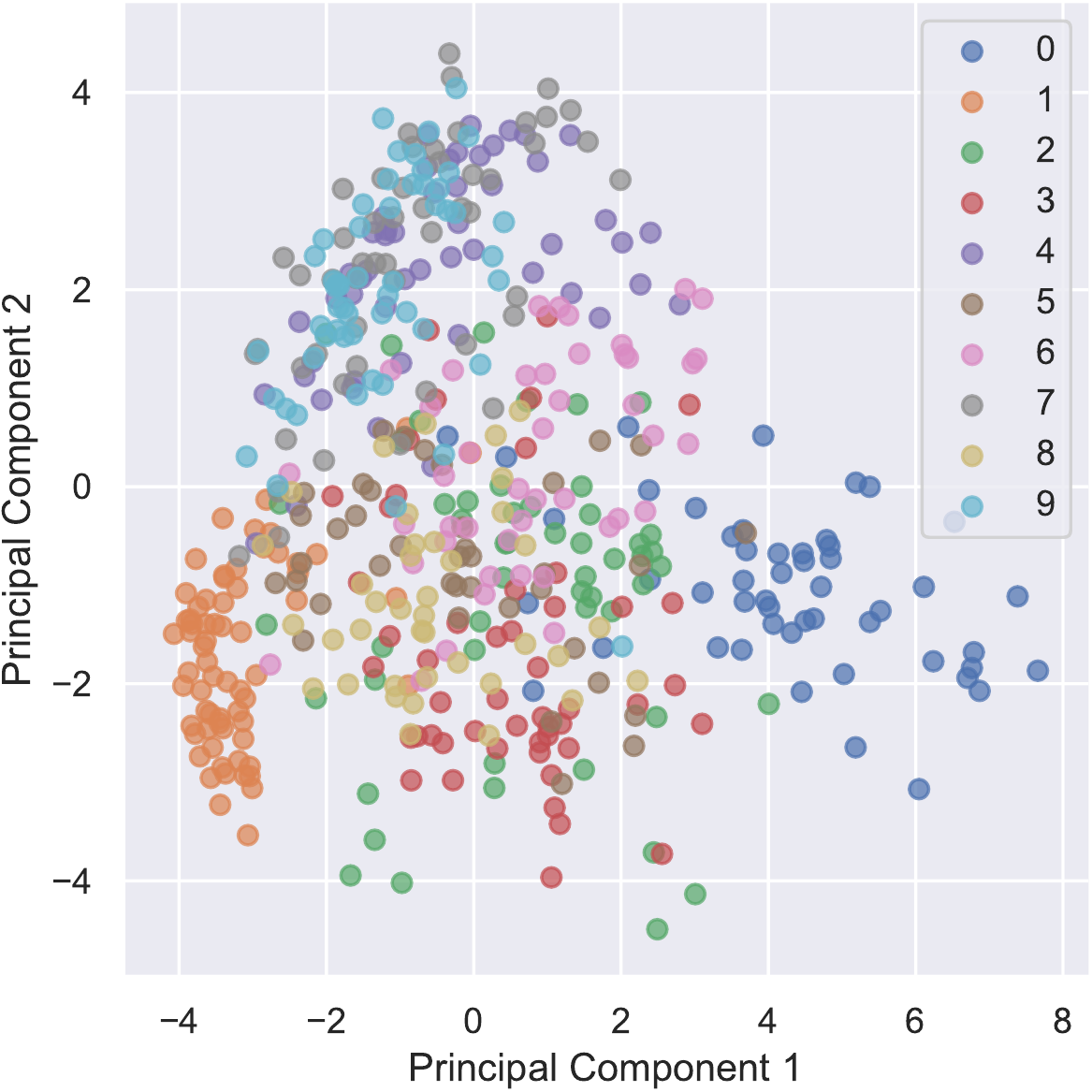}\label{fig:pca_mnist}}

\caption{Principal component analyses of datasets showing class correlations between 2 components.}
\label{fig:pca_analysis}
\end{figure*}

For a set of \ac{ML} datasets, each hyperparameter is iterated for its allowable values on the framework. In each iteration a training, followed by an inference experiment are carried out to study the tradeoffs between accuracy, performance and energy. In our energy-frugality investigations, we use three characteristically different datasets as follows (also see~\cref{fig:pca_analysis}). \textit{Iris}\footnote{Iris: https://tinyurl.com/y5w8zkvj} is a small flower detection dataset with 16 booleanized digits; the dataset has 3 output classes which are evenly distributed in the 150 datapoints but with high correlations between two (\textit{versicolor} and \textit{virginica}) output classes. \textit{Breast Cancer}\footnote{Breast Cancer: https://tinyurl.com/gl3yhzb} is a diagnostic dataset with 300 booleanized digits; the dataset has 2 output classes (\textit{malignant} and \textit{benign}) with a $9$:$7$ bias towards \textit{benign} in the 569 datapoints with minor correlations between them. \textit{MNIST}\footnote{MNIST: https://tinyurl.com/cpbyrhs} is a handwritten digit recognition dataset with 784 booleanized digits; the dataset has 10 output classes without any particular bias between them. However, it features high correlations between some digits, \eg{} between $5$ and $6$, and between $1$ and $7$.

\subsection{Hyperparameter Search and Design Exploration} \label{sec:design:search}

\begin{figure*}[htbp]
    \centering 
    
\subfloat[\textit{Iris}: $s$=1.2]{\includegraphics[width=0.32\textwidth]{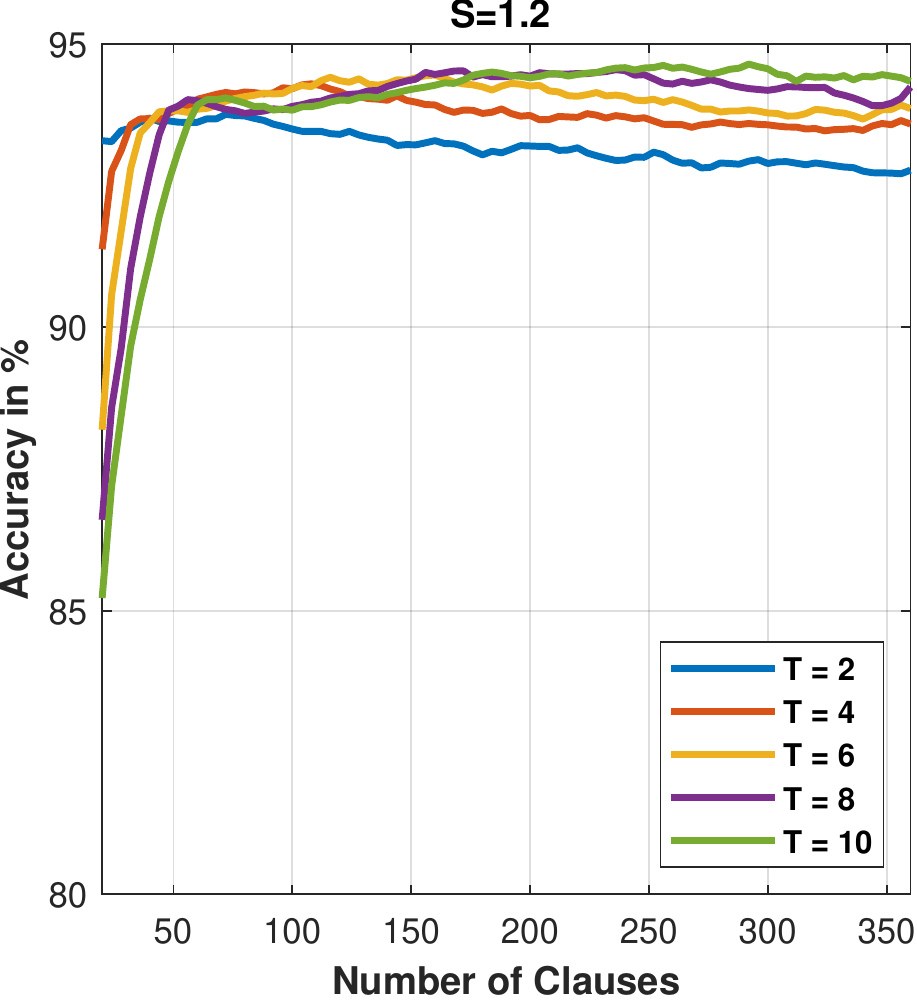}\label{fig:a}}
\subfloat[\textit{Breast Cancer}: $s$=1.1]{\includegraphics[width=0.32\textwidth]{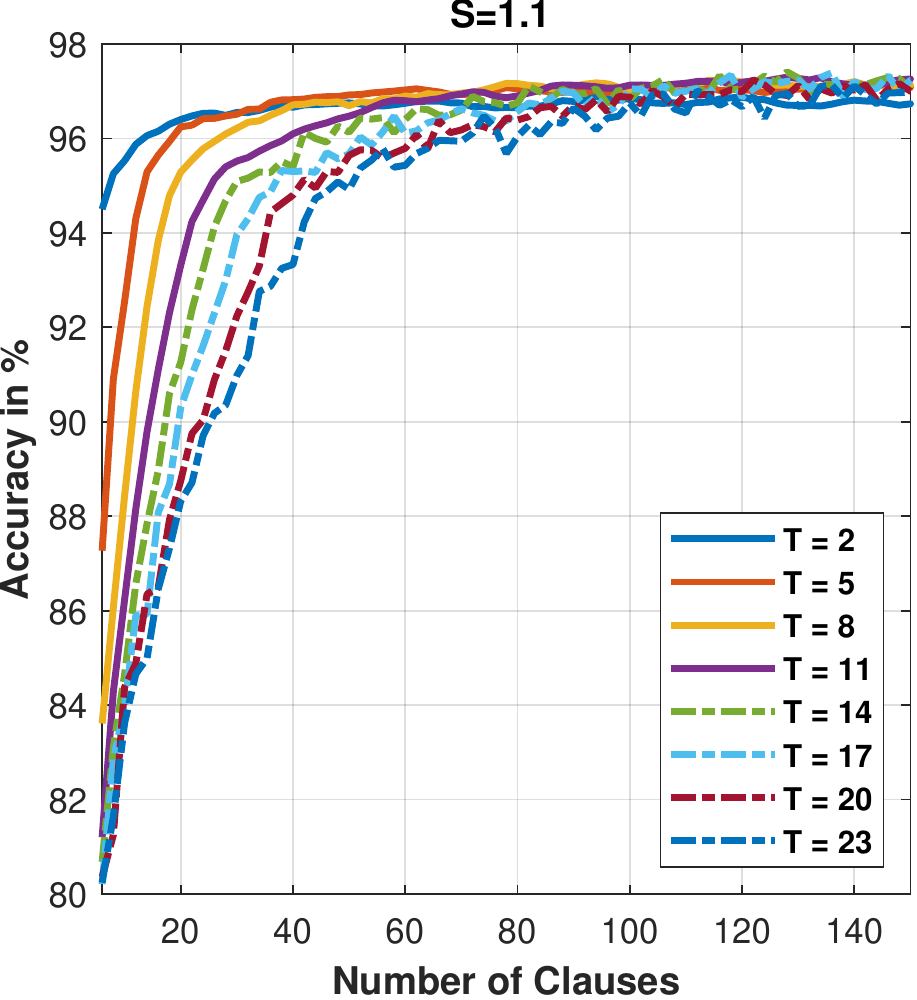}\label{fig:b}}
\subfloat[\textit{MNIST}: $s$=1.9]{\includegraphics[width=0.32\textwidth]{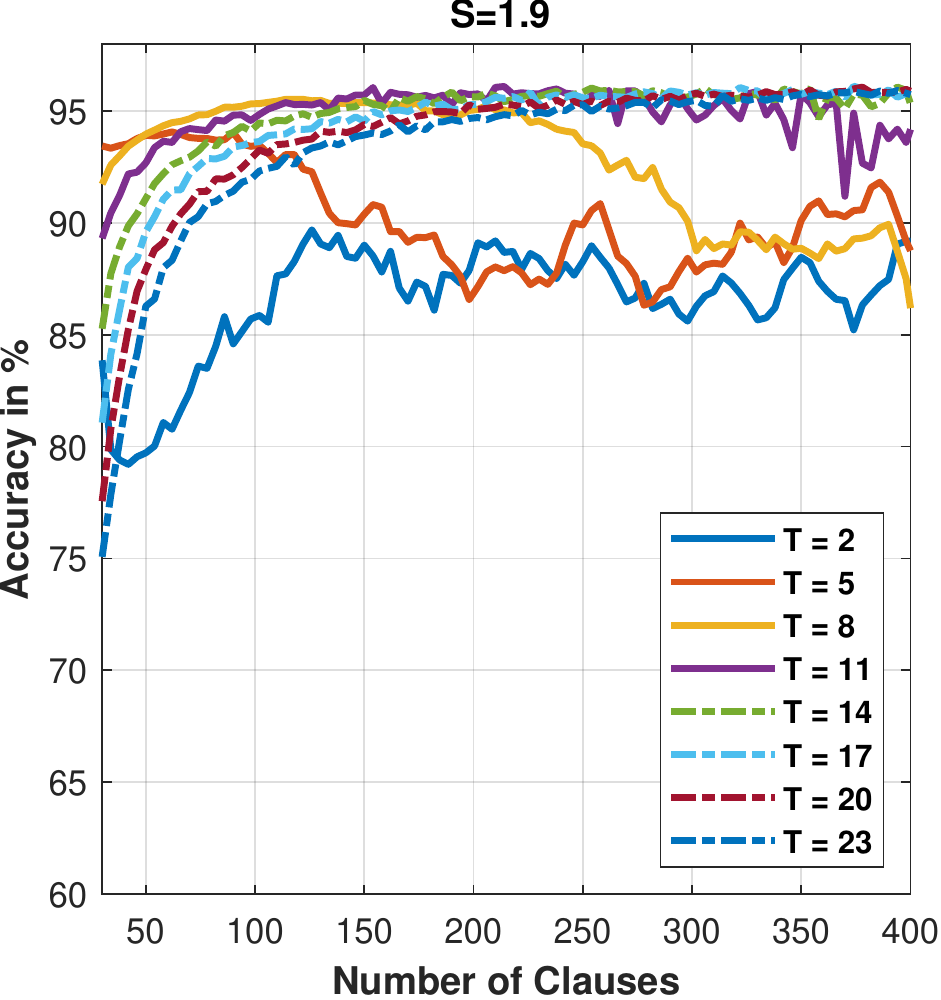}\label{fig:c}}


\medskip

\subfloat[\textit{Iris}: $s$=10]{\includegraphics[width=0.32\textwidth]{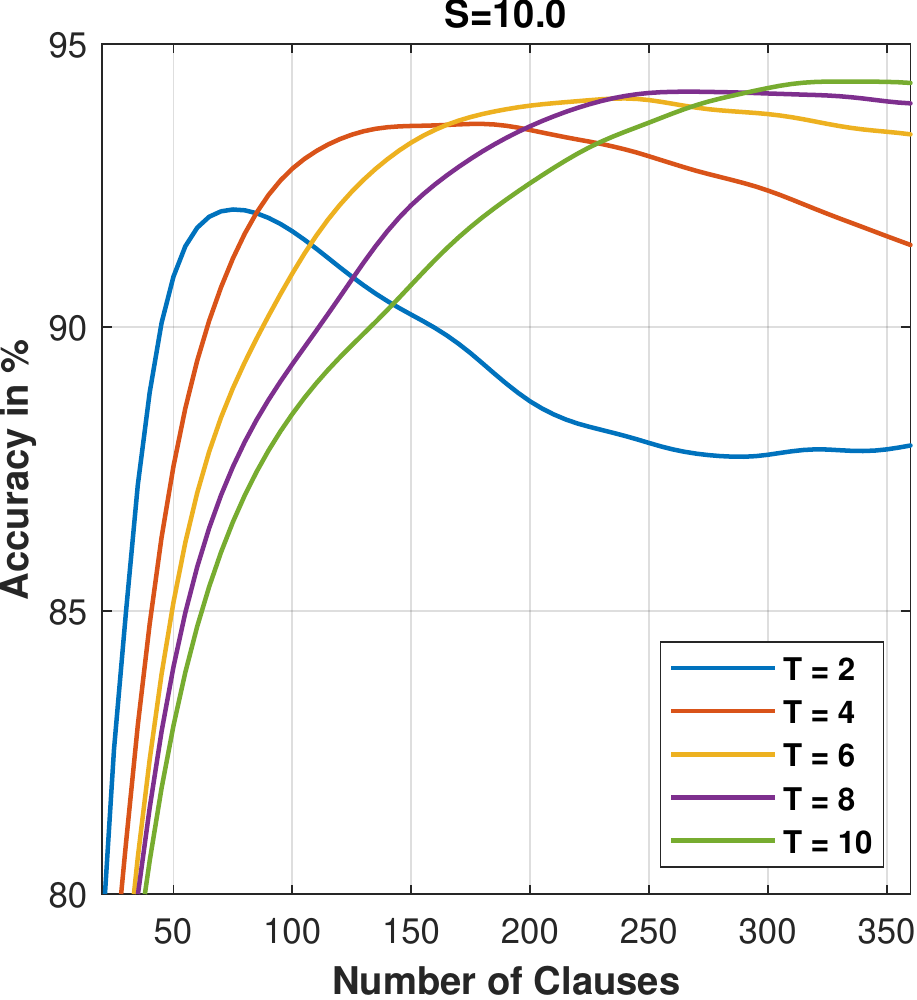}\label{fig:d}}
\subfloat[\textit{Breast Cancer}: $s$=5]{\includegraphics[width=0.32\textwidth]{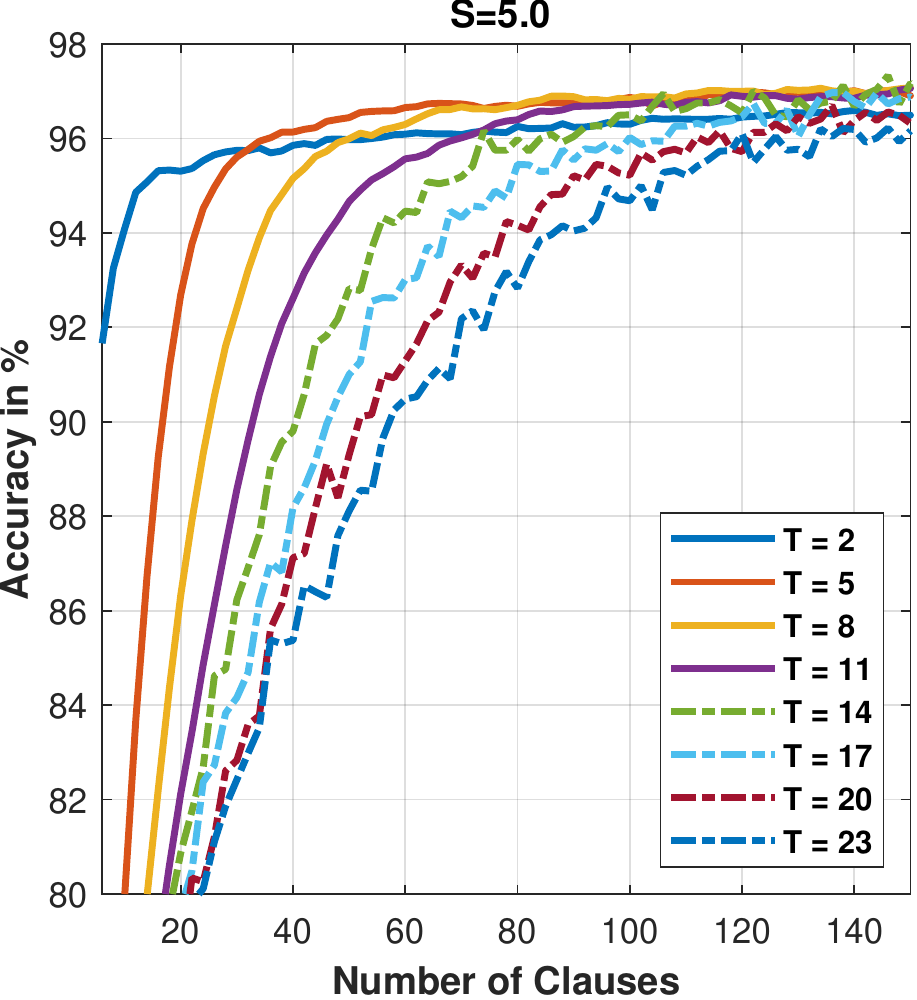}\label{fig:e}}
\subfloat[\textit{MNIST}: $s$=20]{\includegraphics[width=0.32\textwidth]{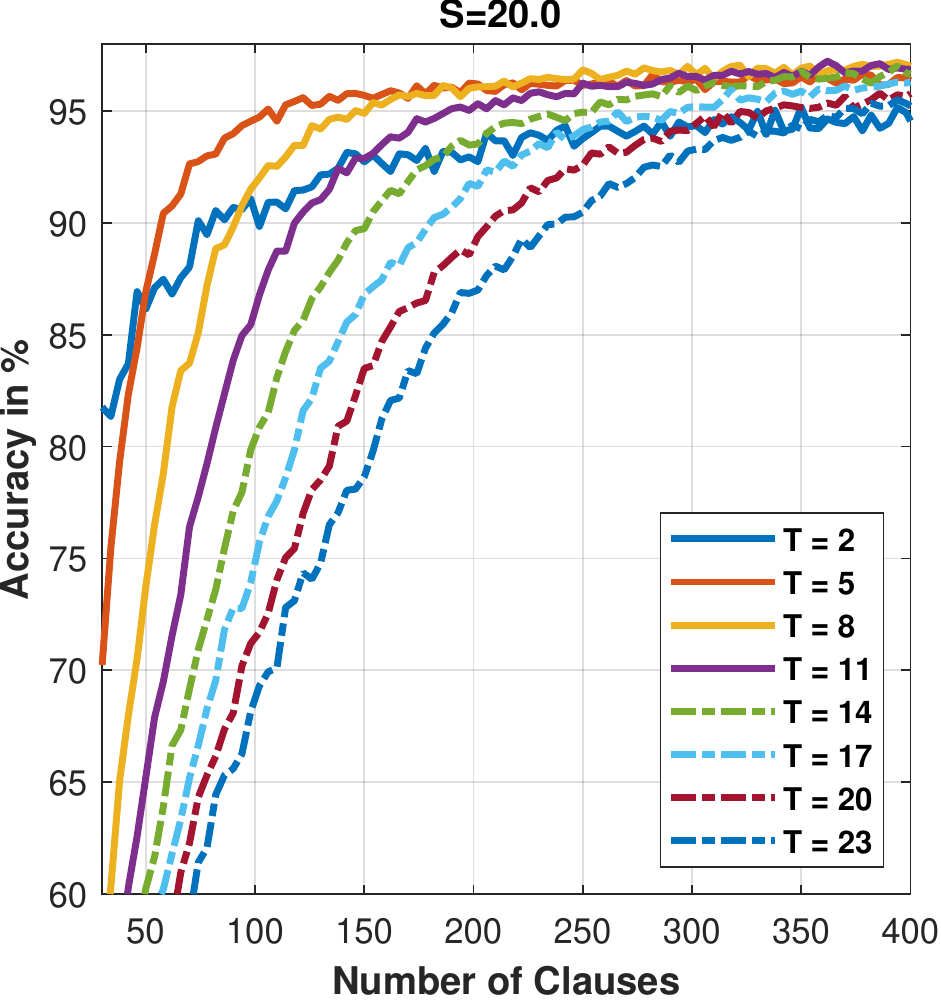}\label{fig:f}}

\caption{The impact of hyperparameters on the \ac{TM} inference accuracy.}
\label{fig:hyperparameter_search}
\end{figure*}

A number of hyperparameter search experiments were carried out using the framework (\cref{fig:design}). \Cref{fig:hyperparameter_search} illustrates the impact of hyperparameter $s$ on the inference accuracy with varying number of clauses ($\Nclauses$) and learning threshold ($T$) values for 3 datasets. Each inference experiment was run with 80\%-20\% split between training and inference for 100 training epochs, which are large enough for learning convergence. This was repeated for 300 times to provide a stable mean of accuracy.

\begin{figure}
		\centering
         \includegraphics[width=.45\linewidth]{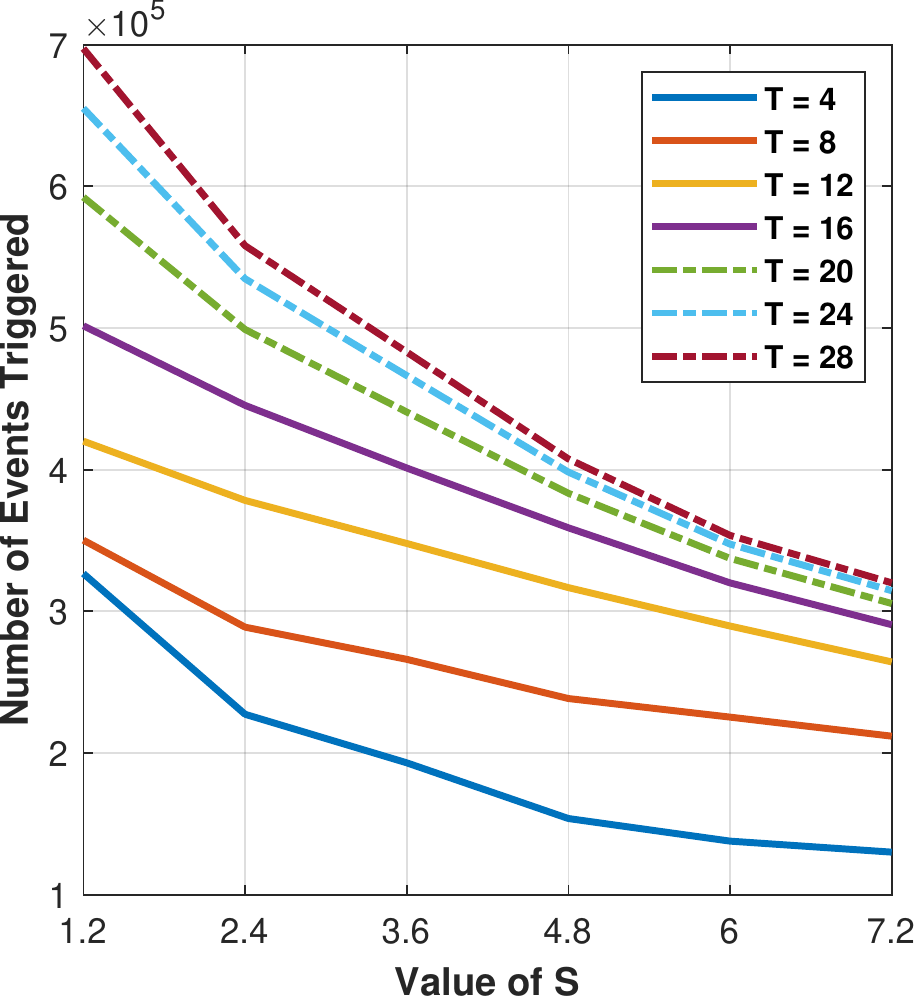}
		\caption{Number of learning events generated by different $T$ and $s$ values on the Iris dataset.}
		\label{fig:iris_events}
\end{figure}

\Cref{fig:hyperparameter_search}(a) and \Cref{fig:hyperparameter_search}(d) show the impact of lower and higher $s$ values ($s$=1.2 and $s$=10) for the \textit{Iris} dataset. As can be seen, smaller $s$ value provides comparable accuracy with lower number of clauses (\eg{} 50 clauses provide the same accuracy for $s$=1.2 as opposed to 150 clauses when $s$=10). This is because lower $s$ value enables more penalty\slash reward events and less no action events per automaton (see~\cref{tab:feedback} and~\cref{fig:iris_events}). In the case of higher $s$ the impact of different $T$ values is remarkably more visible. This is an effect of lower $T$ values enabling less number of randomly selected concurrent \ac{TA} updates in the same reinforcement cycle (\cref{fig:iris_events}). For both experiments, when the number of clauses is substantially large (\eg{} $> 300$), maximum accuracy can be achieved at higher $T$ values. Due to sporadic \ac{TA} reinforcements dispersed between the clauses, lower $T$ values show over-fitting trends between the \textit{verisicolor} and \textit{virginica} output classes, which have high correlations (\cref{fig:pca_analysis}(a)).

The over-fitting problem is however less dominant in the \textit{Breast Cancer} dataset as shown in \Cref{fig:hyperparameter_search}(b) and \Cref{fig:hyperparameter_search}(e) as there is less correlations between the output classes. As such, lower $T$ and $s$ values contribute to higher accuracy even at a lower number of clauses (\cref{fig:pca_analysis}(b)). For example, despite having a larger number of datapoints this dataset can achieve more than 96\% accuracy with only 60 clauses. 

The hyperparameter search experiments in \textit{MNIST} provide similar insights as in \textit{Iris} as lower $T$ values persistently suffer from over-fitting (\Cref{fig:hyperparameter_search}(c) and \Cref{fig:hyperparameter_search}(f)). With high correlations between classes in the dataset, more concurrency and stochasticity are essential through higher $T$ values. 

\begin{table}[htbp]
\centering
\small{
		\caption{Careful prodigality offers faster learning, although with more number of reinforcement events per learning epoch. Note here, the number of reinforcements through all feedback types are almost equally inflated when $s$ is higher and $T$ is lower as more learning epochs are necessary for comparable accuracy.}
		\label{tab:iris_events}
		\centering
		\begin{tabular}{c||c|c|c|c|c|c}
		
     \textbf{Setup} &  \textbf{s} &  \textbf{T} &  \textbf{Epoch} &  \multicolumn{1}{m{2.8cm}|}{\textbf{Reward Type I}}&  \multicolumn{1}{m{2.8cm}|}{ \textbf{Penalty Type I}}&  \multicolumn{1}{m{2.8cm}}{\textbf{Penalty Type II}}  \\           
     \hline
     \hline
     1 & 1.2 & 20 & 3   & 62,086 & 52,258 & 15,909\\  
     2 & 18  & 4  & 100 & 102,510 & 137,491 & 36,679\\\hline
    \end{tabular}
    }
\end{table}

\cref{fig:iris_events} shows the number of penalty\slash reward reinforcement events generated by the \textit{Iris} dataset for different ($s$, $T$) pairs. The number of clauses is fixed at 90 and the inference experiments were executed over 30 epochs. As expected, low $T$ produces significantly less number of events when compared with higher $T$ for a given $s$. Higher $s$ value reduces the number of events considerably, but introduces further instability in the learning (\cref{fig:hyperparameter_search}(d)). Although in terms of energy consumption this is more rewarding, this can provide inferior inference performance due to over-fitting issues described earlier (see \cref{fig:hyperparameter_search}(a)). Higher $T$ with lower $s$ values produce more events and provide better learning efficacy (see~\cref{fig:hyperparameter_search}(a) and~\cref{fig:hyperparameter_search}(d)) at the cost of higher energy consumption. These are examples of conflicting tradeoffs, where prodigality must be carefully designed to ensure a balance between energy and learning efficacy. \Cref{tab:iris_events} demonstrates that the prodigal allocation (lower $s$ with higher $T$) exploits the stochastic diversity between \ac{TA} better and achieves $>$93\% accuracy much faster (in 3 epochs) when compared with the learning efficacy unaware and energy-frugal solution (higher $s$ with lower $T$).

\Cref{tab:energy_accuracy} shows the optimized hyperparemeters that offer the best learning efficacy and energy-frugality objectives. To generate the energy figures, we used low-level hardware design experiments in \pnoun{Cadence Innovus} using scaled-down datasets (\eg{} \textit{Iris}). A \SI{65}{\nm} low-power technology node based design was synthesized with necessary peripherals and \ac{IO} to produce normalized energy in terms of the energy consumed per atomic data operation~\cite{Wheeldon2020a}. These were then scaled up for the optimized architectural allocations in each dataset to estimate the energy required per inference datapoint (\ie{} the number of concurrent booleanized literals).
\begin{table}[ht]
  \centering
\small{
  \caption{%
     Optimized hyperparameters, learning accuracy and energy consumption for datasets.
  }\label{tab:energy_accuracy}
\begin{center}
\begin{tabular}{c||c|c|c|p{0.5in}|p{0.5in}}
 \textbf{Dataset} &  \textbf{Clauses} &  $T$ & $s$ &  \textbf{Testing Accuracy} & \textbf{Energy\slash datapoint}\\           \hline
         \hline
 \textit{Iris} & 90 & 4 & 1.2 &94.1\% & 68.8 \textit{pJ}\\  
 \textit{Breast Cancer}  & 60 & 5 & 1.1 & 97.0\% & 574 \textit{pJ}\\  
 \textit{MNIST} & 100 & 8 & 1.9 & 95.1\% & 12.5 \textit{nJ}\\\hline
\end{tabular}
\end{center}
}
\end{table}

\subsection{Impact of Pseudorandom Number Generation on Accuracy}

Pseudorandom number generation is an important means for ensuring stochastic diversity between clauses in \ac{TM}. 
Combined with the agility in issuing reinforcement actions (by means of parameter $s$), this becomes an effective mechanism to control the prodigality and hence accuracy of a \ac{TM}. In the hardware implementation of \ac{TM}, one \ac{PRNG} is instantiated for each \ac{TA} to allow maximum concurrency in learning. In its current form, the software \ac{TM} produces random numbers using a $64$-bit \ac{PCG}. Whilst the \ac{PCG} offers great statistical properties, it requires a $64\times 64$-bit multiplication, addition and shifts. These operations are very costly to implement in hardware in terms of area and power.
\begin{figure}[htbp]
\centering

\subfloat[\acs{LFSR} bit width vs \ac{TM} test accuracy.]{\includegraphics[width=0.46\textwidth]{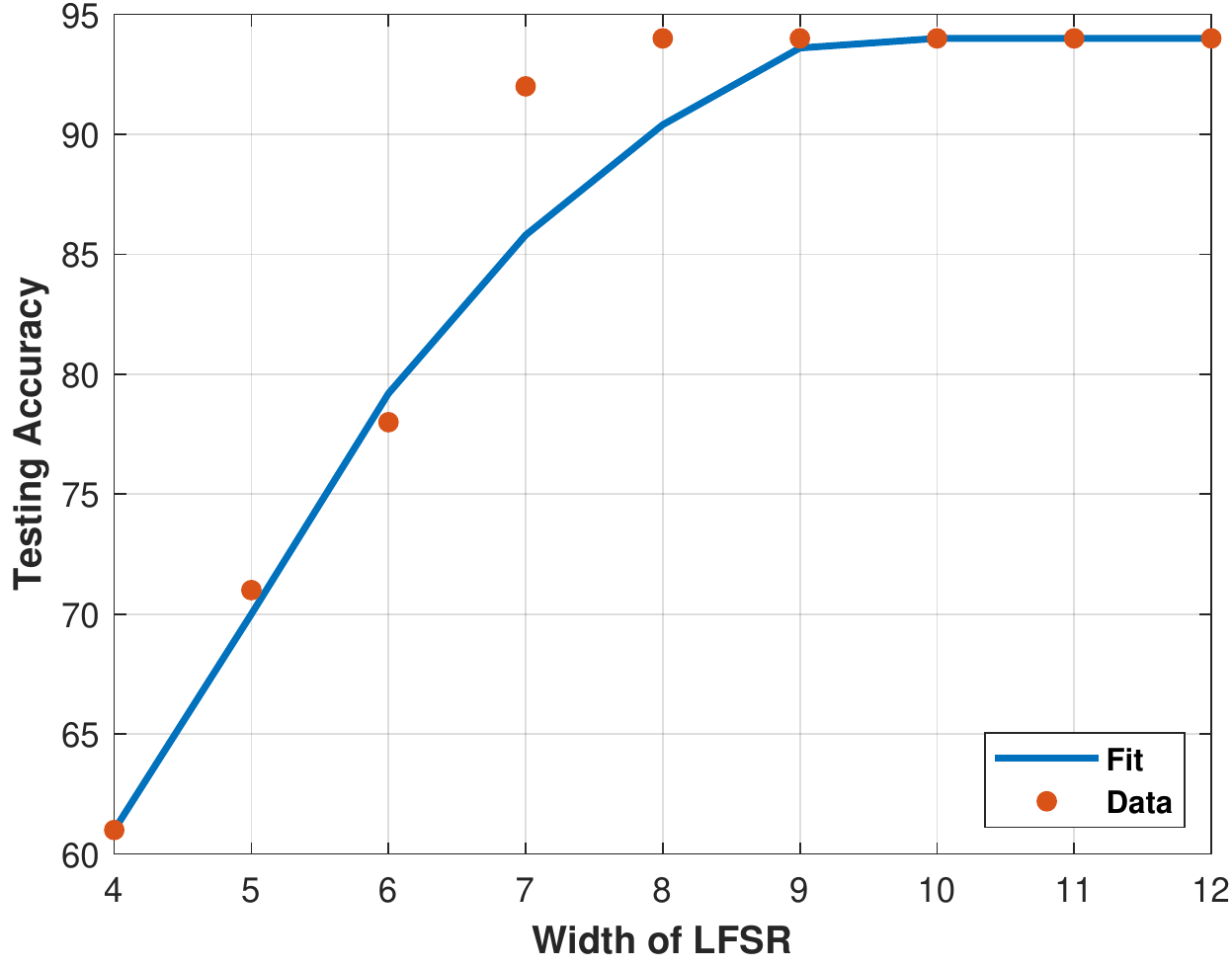}\label{fig:lfsr-vs-accuracy}}
\subfloat[$s$ parameter vs test accuracy for the $6$-bit \acs{LFSR}.]{\includegraphics[width=0.46\textwidth]{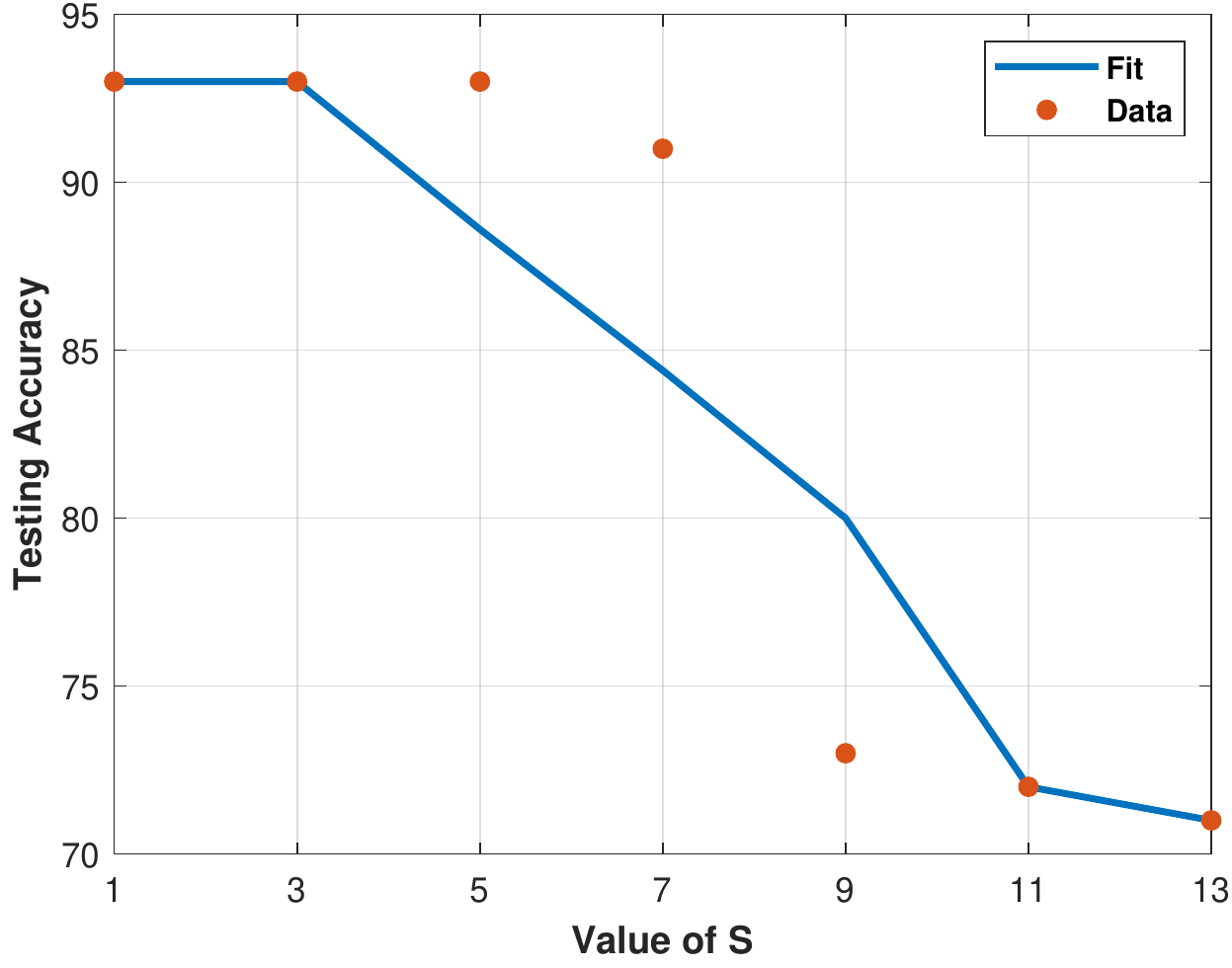}\label{fig:s-vs-accuracy}}

    \caption{The impact of LFSR allocations on learning efficacy.}
\end{figure}

In the \ac{TM} we are not concerned with the unpredictability properties of the \ac{PRNG}, but only the stochastic diversity between the clauses. So as an alternative to the \ac{PCG}, we instead use \acp{LFSR} which require only a shift register and \gate{xor} operations. The area and power of \iac{LFSR} scales with its bit width, therefore minimization of bit width is paramount for energy frugality.

In order to assess the impact of stochasticity on the \ac{TM}, we first find the $(\Nclauses, T, s)$ parameter combination giving the highest accuracy on the Iris dataset, using the original \ac{PCG} \ac{PRNG} method, $50$ epochs and $1000$ ensembles: $(140, 11, 10)$.
Keeping these parameters constant, in \cref{fig:lfsr-vs-accuracy} we investigate how differently sized \acp{LFSR} compare in accuracy with \ac{PCG}. The $8$-bit \ac{LFSR} maintains similar accuracy to the \ac{PCG}---in fact, any \ac{LFSR} size from $8$ to $32$ bits exhibit accuracy within the margin of error. With a $7$-bit \ac{LFSR} there is a small but noticeable drop in accuracy. Below $7$-bit width we see a huge and unacceptable loss in accuracy as the diversity in clause learning drops.

\Cref{fig:s-vs-accuracy} shows that, for this dataset and \ac{TM} configuration, it is possible to regain some accuracy lost by the $7$-bit \ac{LFSR} by reducing the value of the $s$ parameter, encouraging more learning steps to take place within the same number of epochs. It is noticed that this loss in accuracy cannot be reclaimed by either increasing the number of clauses or varying the $T$ parameter. In the former case, clauses are simply duplicated and do not add any extra information to the learning process. This means that the optimal \ac{TM} specification using \ac{PCG} is also the optimal specification for \acp{LFSR} of width $8$-bits or greater with the Iris dataset. It is not clear how datasets with higher dimensionalities will be affected by the precision of \ac{LFSR}; we aim to study this in the future.

	\section{Explainability and Dependability using Reachability Analysis} \label{sec:reachability}

Modeling learning capability is vital for understanding explainability~\cite{bhatt2020explainable}. A crucial component of this capability is reachability analysis. Reachability is traditionally defined as a process of exploring the set of states that a (usually discrete-event) system can visit while performing a set of permitted actions. Often this process aims to check for and prove certain properties of the system. In our research, we define reachability slightly more specifically, as the property of the system that allows it to navigate through the finite state-space produced by the composition of finite-state automata, namely \ac{TA}s. This property is crucial for the hardware to generate the intended and bounded outputs by relating them to sequences of the input data points. 

To investigate reachability of the \ac{AI} hardware using the principles of learning automata (Figure~\ref{fig:tsetlin}), a key hardware block is the team of \ac{TA} within the reinforcement component. 
\begin{figure}[htbp]
    \centering
    \includegraphics[width=0.6\columnwidth]{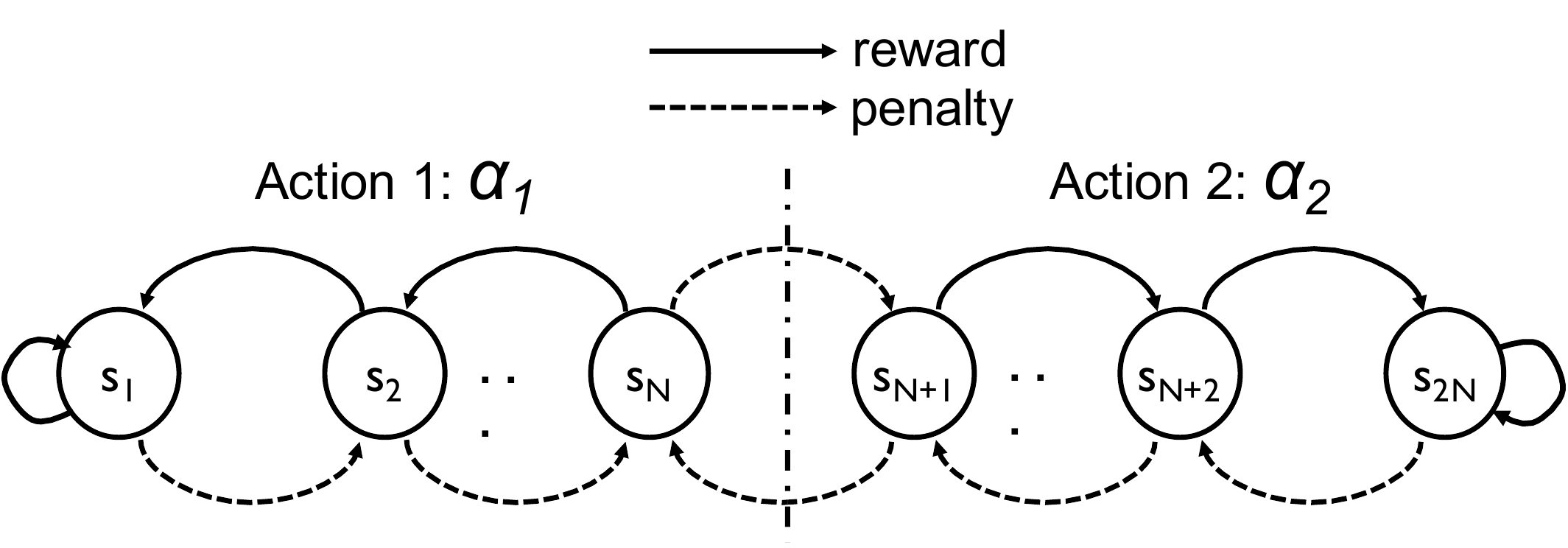}
    \caption{A Tsetlin automaton for 2-action environment with $2N$ states}
    \label{fig:states}
\end{figure}
As described earlier, the overall operational cycle in reinforcement involves the work of both sequential (\ac{TA}s) and combinational parts (clauses, classifiers and feedback). As input data sequences are applied, the whole system evolves in the TA state-space and eventually reaches the subset of states (trained states) where the system can perform its most advantageous classification decisions. The latter property, convergence to the stable trained state, is crucial for the accuracy and efficiency of the \ac{TM} in terms of performance and energy. Besides, reachability becomes a measure of explainability because the trajectories of states through which the system converges can be easily traced. 

\begin{figure*}[htbp]
\centering
\subfloat[Reinforcing 4 TA with datapoints ($X_0,X_1$): (0, 0)]{\includegraphics[width=0.75\textwidth]{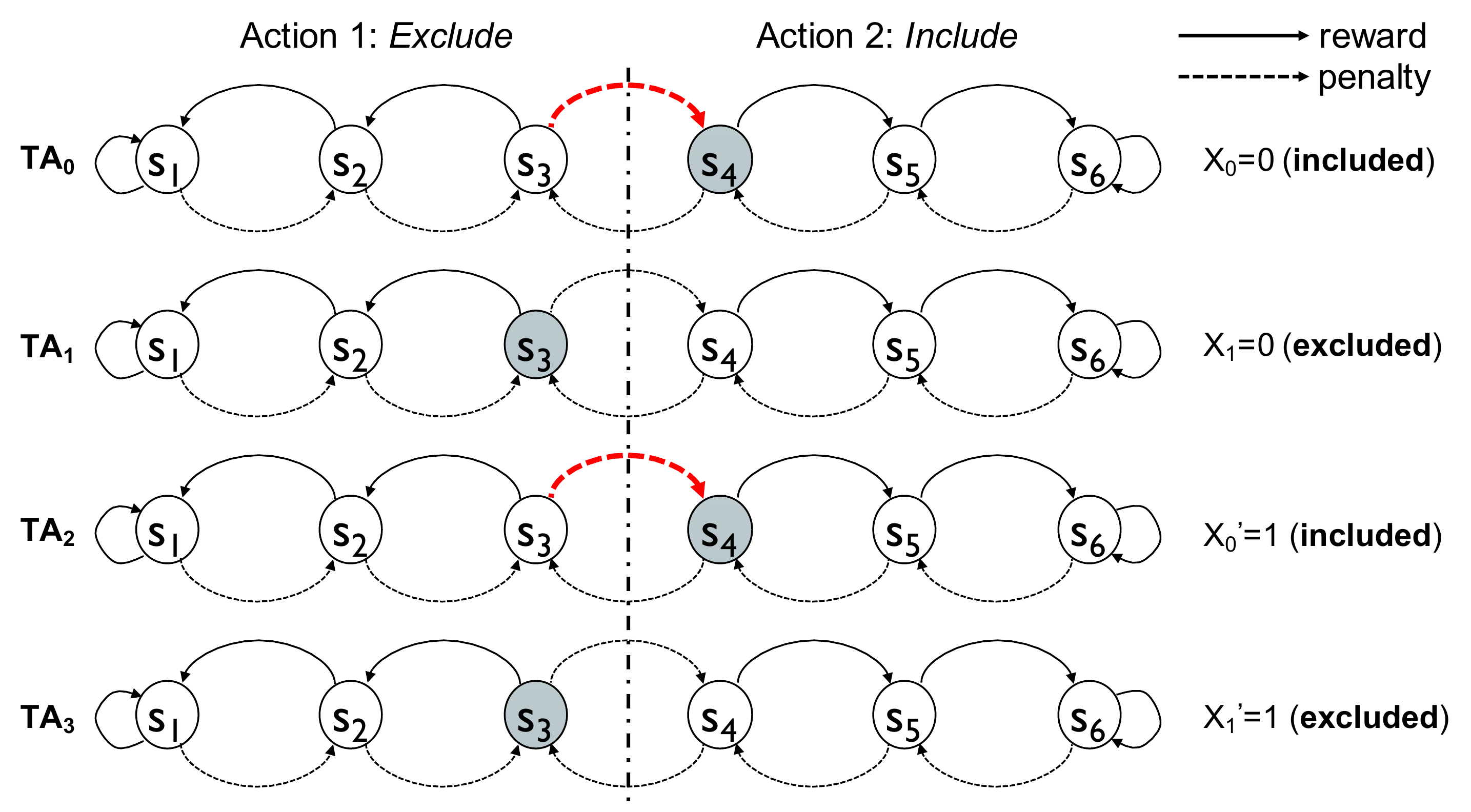}}\\
\subfloat[Reinforcing 4 TA with datapoints ($X_0,X_1$): (0, 1)]{\includegraphics[width=0.75\textwidth]{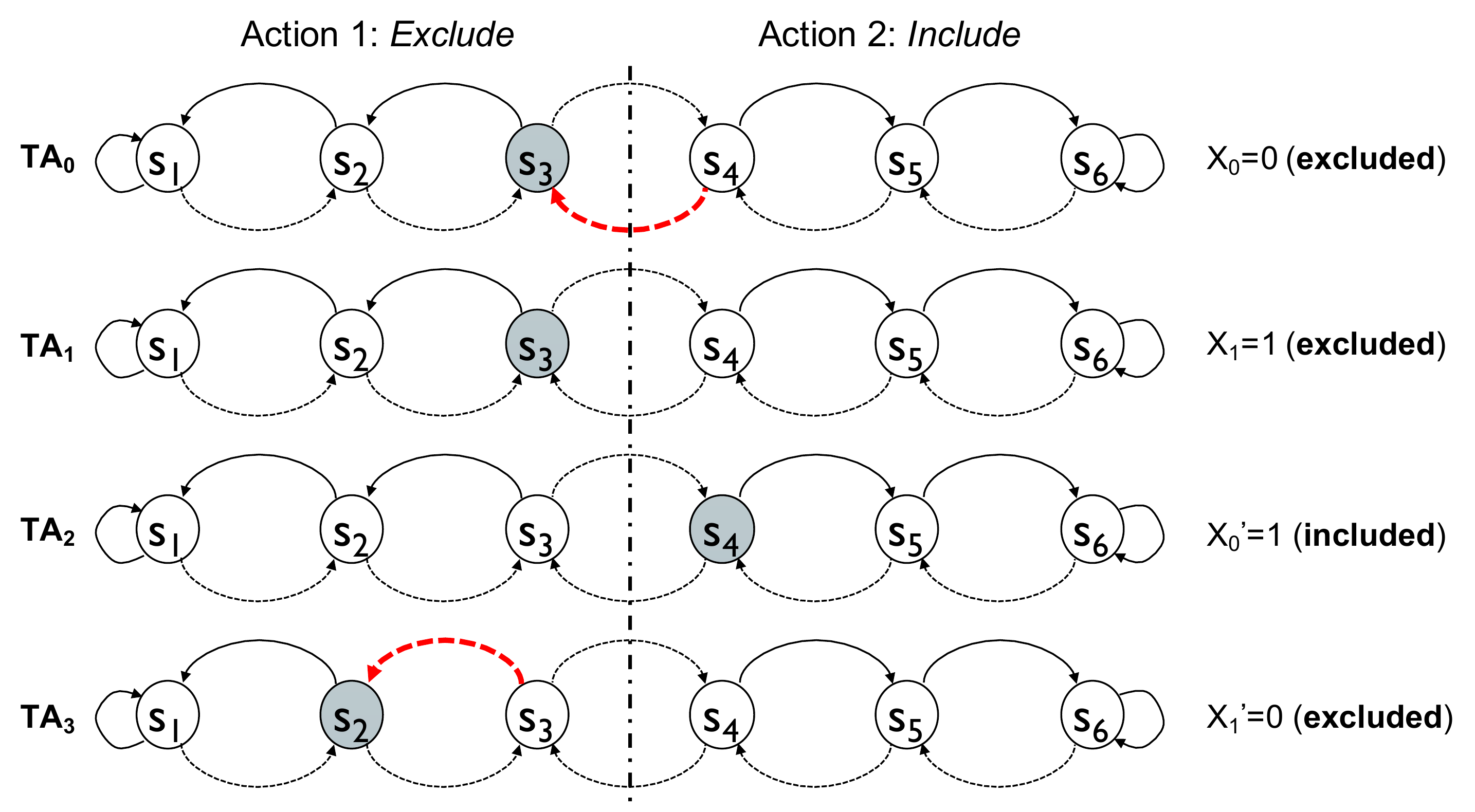}}\\
\subfloat[TA states After 11 reinforcement steps (i.e. 44 datapoints)]{\includegraphics[width=0.75\textwidth]{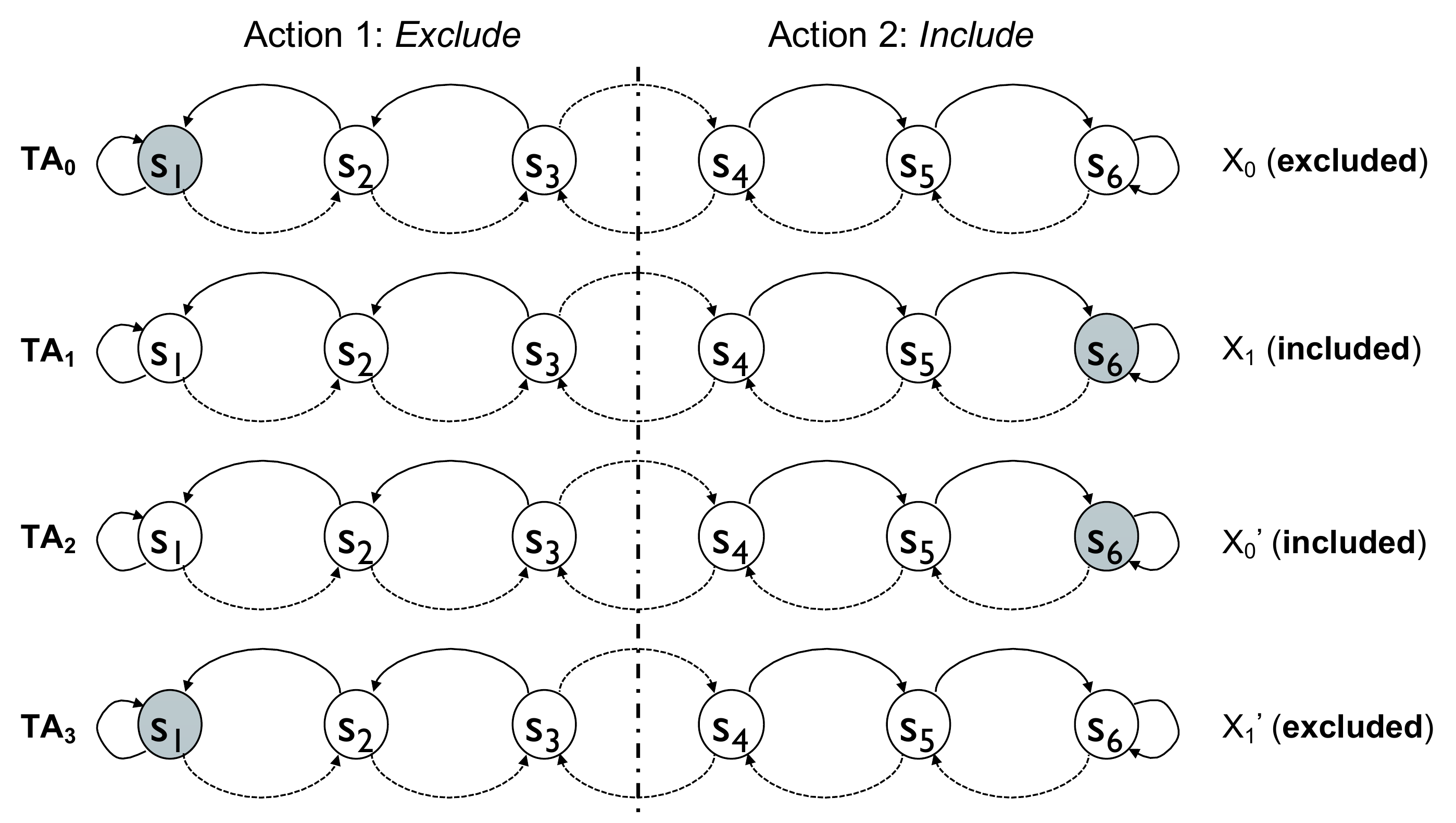}}
\caption{An illustrative example of TA state changes in a 2-input binary XOR}
\label{fig:ta-states}
\end{figure*}
Figure~\ref{fig:states} shows a high-level state transition diagram of each automaton with $2N$ internal states in a 2-action environment. We denote the TA states as $\textbf{S}$=$\{s_1, s_2, \dots s_n \dots s_{2N}\}$, where $s_n$ is the $n$-th state. Each automaton initially starts with a random state near the action boundary, \eg{} $\phi_{init}$=$s_{N}$ or $s_{N+1}$. This allows for the \ac{TA} to make less number of state transitions to reinforce an action. After each reinforcement step, a reward is used to strengthen an action or a penalty is used to weaken the automaton confidence in performing the current action~\cite{Granmo2018}. Since state transitions take place in discrete single steps, $s_n$ is the \ac{TA} state resulting from a transition from either of $s_{n-1}$ or $s_{n+1}$. For a given state of $s_n$, the action performed by the automaton is given as:
\begin{equation}
    G(s_n) = \begin{cases}
      \alpha_1; & \text{if $1 \leq n \leq N$}\\
      \alpha_2; & \text{if $(N+1) \leq n \leq 2N$}
    \end{cases} 
\end{equation}

To demonstrate the number of reinforcement steps needed to fully converge to the final state as well as the corresponding action, we consider an automaton with $2N=6$ internal states and 2 actions. The state transition equations of all automaton states are given as below: 
\begin{eqnarray}
s_1 = (s_1\ \text{AND}\ R) + (s_2\ \text{AND}\ R);\\
s_2 = (s_1\ \text{AND}\ P) + (s_3\ \text{AND}\ R);\\
s_3 = (s_2\ \text{AND}\ P) + (s_4\ \text{AND}\ P);\\
s_4 = (s_3\ \text{AND}\ P) + (s_5\ \text{AND}\ P);\\
s_5 = (s_4\ \text{AND}\ R) + (s_6\ \text{AND}\ P);\\
s_6 = (s_5\ \text{AND}\ R) + (s_6\ \text{AND}\ R),
\end{eqnarray}
where R and P are the reward and penalty signals generated by the state update circuit depending on the randomized reinforcement trajectory (\cref{tab:feedback}). From Eqns. (2)-(7), given the random $\phi_{init}$ of either $s_3$ or $s_4$, the automaton needs minimum 3 or 4 reinforcement steps. In Section~\ref{sec:game_theoretic}, we provide a game theoretic analysis of state convergence using Nash equilibria for a binary XOR example. 


The states of the whole \ac{TM} are formed as Cartesian products of the states of individual \ac{TA}s. 
To illustrate how bounded \ac{TA} state transitions contribute to reachable learning formulation in the \ac{TM} algorithm, we simulate a 2-input XOR using a SystemC description of the same. The inputs and their complements constitute 4 literals and as such 4 \ac{TA} are used in each clause. Each automaton consists of 6 states as exemplified above. A total of 4 clauses are used in the inference circuit, of which 2 are positive clauses and 2 are negative clauses into the majority voting (i.e. classification) circuit. Figure~\ref{fig:ta-states} shows the internal states of 4 \ac{TA}, defining one clause output only.

The state transitions in an epoch correspond to 4 datapoints (which are the set of literals), but only 2 are shown. The \ac{TA} start with the same initial states of $s3$. After the first datapoint ($\textbf{X}$:[$X_0, X_1$]=[$0,0$]) reinforcement into literals $l_0$, $l_0'$, $l_1$ and $l_1'$ (including the original booleans and their complements), the clause sees an output of 1 as all \ac{TA} states suggest no inclusion of 0 literals. Overall, this results in a false positive classification and as such 2 penalties in TA$_0$ and TA$_2$ through type II feedback (\cref{tab:feedback}), causing them to transition to $s_4$ (Figure~\ref{fig:ta-states}(a)). After the second datapoint ($\textbf{X}$:[$0,1$]), the clause output is 0 as the TA$_2$ state favors the inclusion of a 0 literal ($X_0^\textbf{'}$). However, due to false negative classification, TA$_0$ is penalized to $s_3$ and TA$_3$ is rewarded to $s_2$ through type I feedback (Figure~\ref{fig:ta-states}(b)). With more datapoints and their associated single-step reinforcements (Eqns.(2)-(7)), the \ac{TA} continue to settle for the states with higher reward probabilities, e.g. $s_1$ and $s_6$ (Figure~\ref{fig:ta-states}(c)). This guarantees convergence during training (also see Section~\ref{sec:game_theoretic}).

The above finite-state reachability analysis shows an important property of the \ac{TM}, where the (integer) vector of states of \ac{TA}s is effectively mapped (contracted) onto the (binary) vector of actions include/exclude. This mapping allows us to define the notion of equivalence between the states of \ac{TA}s, and hence define the conditions for detecting convergence to the trained state as soon as possible, thus improving the efficiency of the system and its performance.  

We continue our reachability analysis in the presence of faults. For this, we introduce fault injection handles in the SystemC model using fault-enabled data types~\cite{shafik2008systemc}. For demonstration purposes, our fault injection campaign includes a stuck-at 1 fault model, applied to the reinforcement part, i.e. \ac{TA}.

\begin{figure}[ht]
    \centering
    \includegraphics[width=0.7\columnwidth]{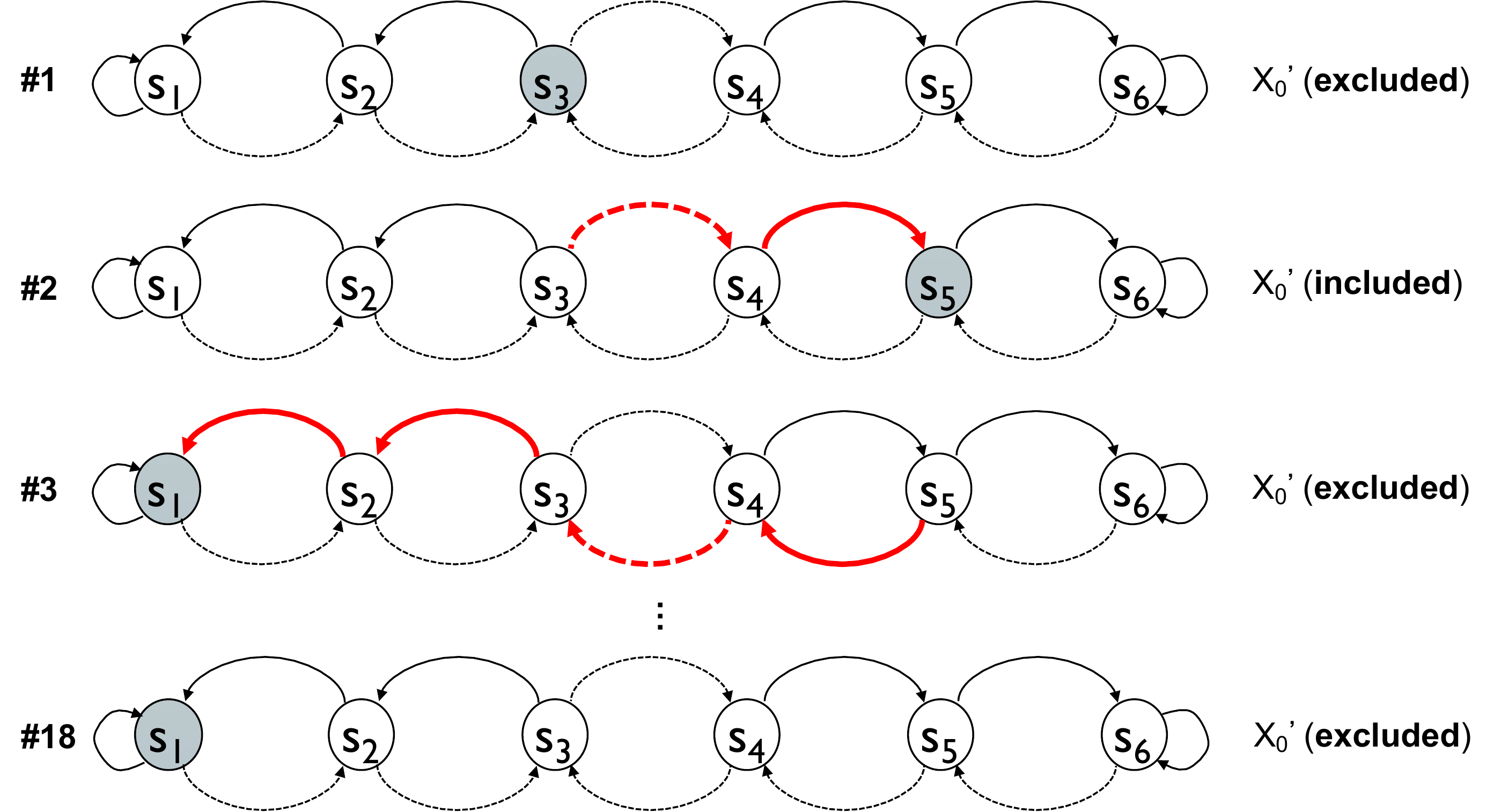}
    \caption{The impact of a stuck-at 1 fault in \ac{TA}$_1$'s state transition and hence learning. Notice how originally included literal is now excluded because of fault in the \ac{TA}.}
    \label{fig:states-fault}
\end{figure}
We inject a stuck-at 1 fault in the least significant bit (i.e. bit position 0) of automaton 1 (i.e. TA$_1$) within the first clause. This is done to observe how this fault can change TA$_1$ state transitions (see Figure~\ref{fig:states-fault}) when compared with the same in Figure~\ref{fig:ta-states}. As can be seen, the automaton assumes an initial state of $s_3$ and does not change the state after the iteration step 1. This is equivalent to a no-action reinforcement of 4 datapoints. In the iteration step 2, the automaton state is penalized towards $s_4$ through an increment operation (i.e. from register value of $011$ to $100$). However, due to the fault the automaton transitions to $s_5$ (i.e. a register value of $101$). After iteration step 3, the automaton is rewarded towards $s6$. However, the faulty automaton state tries to transition to an unreachable state of $s_7$. As the state bounds are protected through a [\textit{modulus} \textit{6} +\textit{1}] operation internally, the automaton changes the state to $s_1$. The automaton retains this state until convergence in all clauses (after 18 epochs). Note that unlike the TA$_1$ state in the first clause of the fault-free \ac{TM} (Figure~\ref{fig:ta-states}), the faulty automaton excludes the associated Boolean literal, $X_0^\textbf{'}$.

From Figure~\ref{fig:states-fault} it is evident that a fault in an automaton can influence its state transitions. For example, the state values of TA$_1$ are constrained to only 3 out of 6 states: $s_1$, $s_3$ and $s_5$, 2 of which are inclined towards the \textit{exclude} action. This results in a maximum achievable accuracy of 75\%. Indeed, defining the relationship between input datapoints and output classes can be challenging with limited state transitions without other means of fault mitigation.
\begin{figure}[htbp]
\centering
\subfloat[Max. accuracy for different number of clauses]{\includegraphics[width = 0.6\columnwidth]{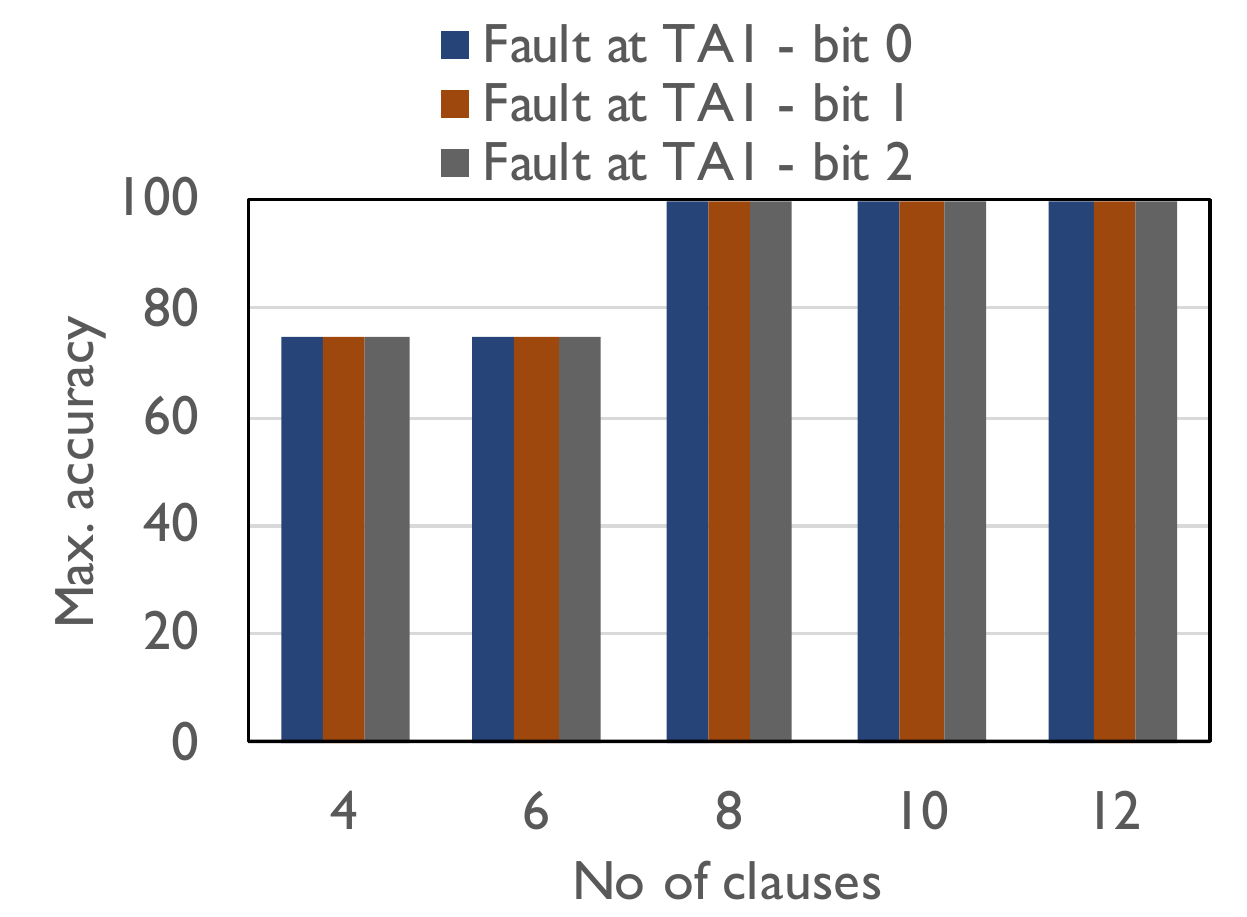}}\\
\subfloat[No of iteration steps for max. accuracy]{\includegraphics[width = 0.6\columnwidth]{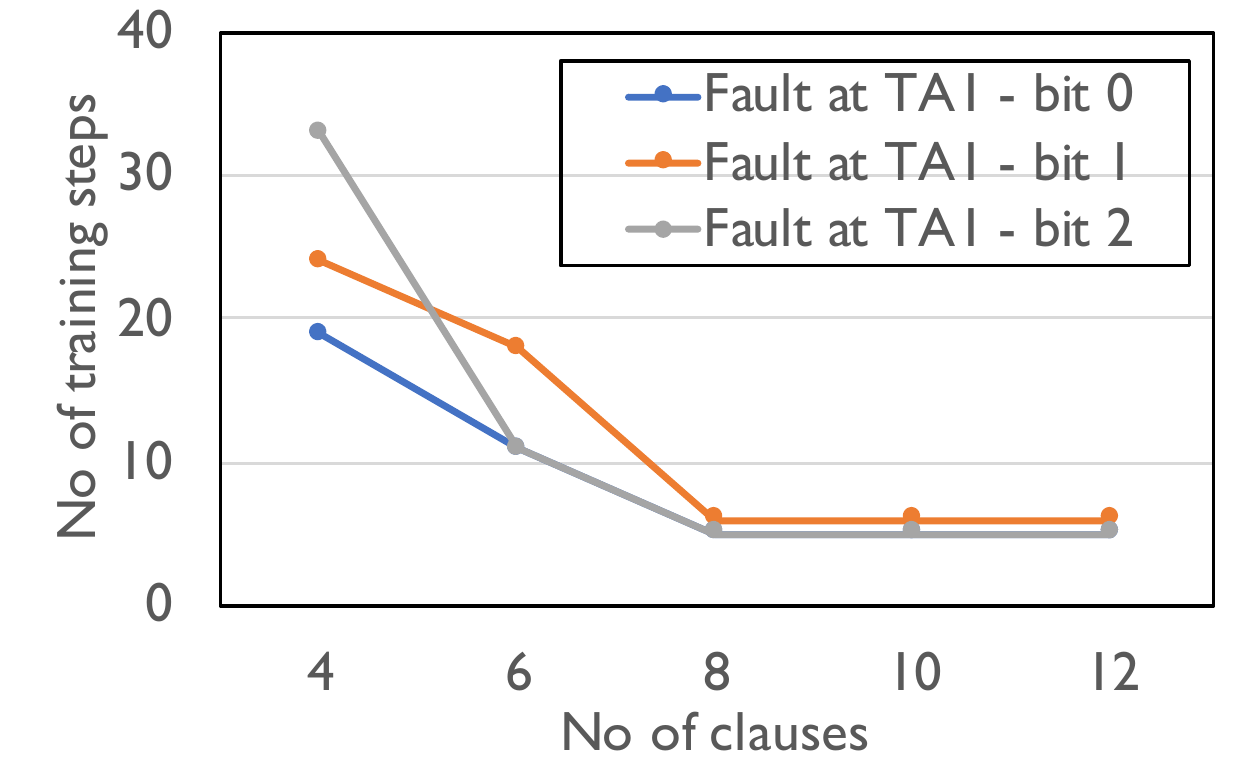}}
\caption{The impact of stuck-at 1 faults in \ac{TA}$_1$ at different bit positions in terms of accuracy and performance; number of clauses are varied to observe how clause redundancy naturally masks the faults.}
\label{fig:dep-clauses}
\end{figure}

Next, we will explore \ac{TM}'s fault masking capability under increased number of clauses, from from 4 to 12 each with 6 automaton states (Figure~\ref{fig:tsetlin}). Figure~\ref{fig:dep-clauses} presents the results in terms of the maximum training accuracy and the corresponding number of iteration steps to convergence. To observe the significance of fault positions, we injected stuck-at 1 faults in different positions of the TA$_1$ register: at bit positions 0, 1 and 2. As expected, when with 8 clauses or more, the training accuracy increases to 100\% (Figure~\ref{fig:dep-clauses}(a)). Provisioning more clauses in \ac{TM} (\ie{} prodigality) allows for increased stochastic variations to ascertain the overall reachability properties~\cite{mathew2014energy}. \ac{TM} features majority voting in the classification circuit and as such mitigation of faults is achieved without any further redundancy control.

\begin{figure}[htbp]
\centering
\subfloat[Max. accuracy for different number of TA states]{\includegraphics[width = 0.6\columnwidth]{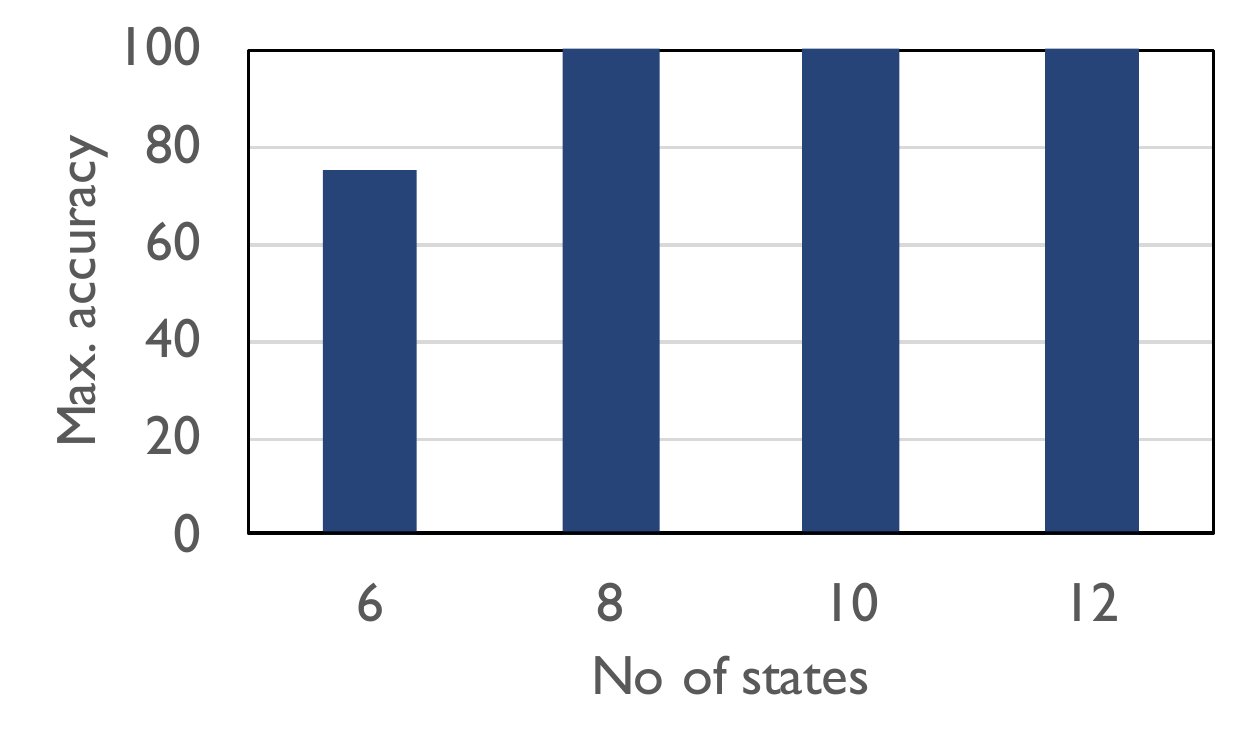}}\\
\subfloat[No of iteration steps for max. accuracy]{\includegraphics[width = 0.6\columnwidth]{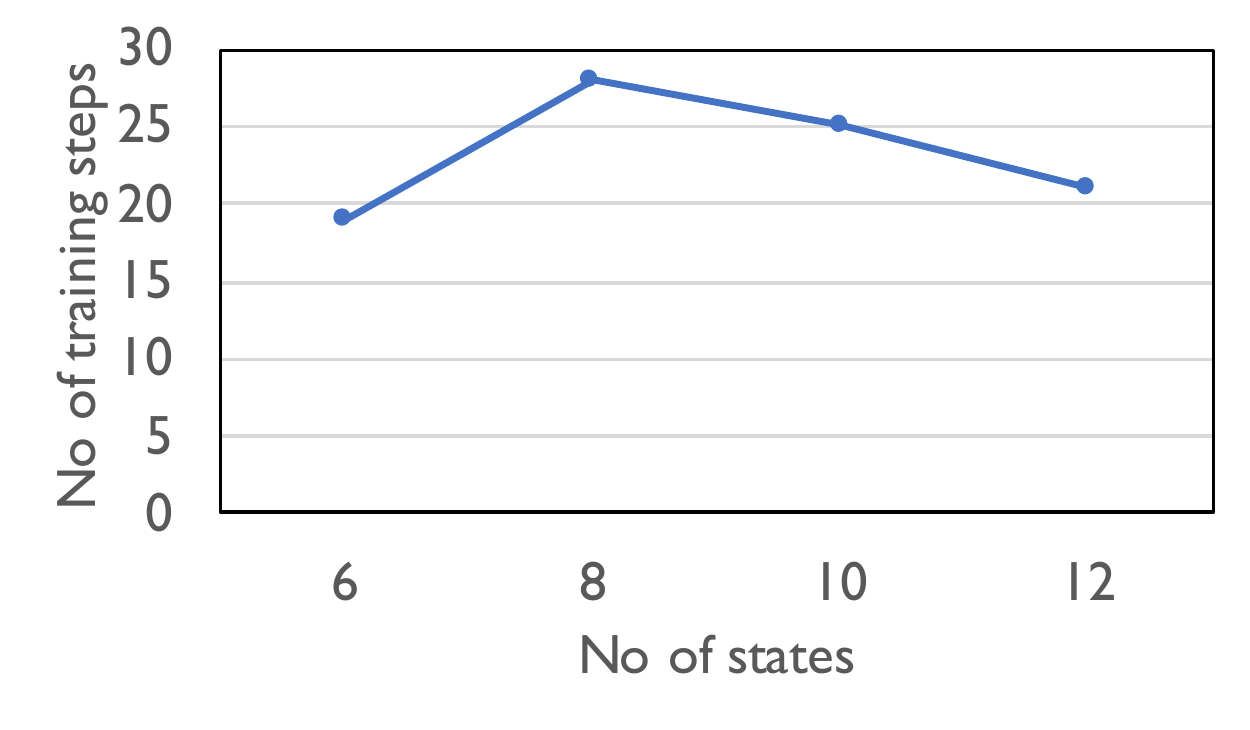}}
\caption{Impact of stuck at faults in $TA_1$ in terms of accuracy and performance with variable number of TA states.}
\label{fig:dep-states}
\end{figure}
The fault positions influence the training times (Figure~\ref{fig:dep-clauses}(b)) as they can constrain the number of state transitions available to an automaton, often with an action bias. Thus, the number of reinforcement steps needed to increase the automaton action confidence is affected. For example, a stuck-at fault in bit position 2 is more challenging to mask as it only allows for the include action states: $s_4$ ($100$), $s_5$ ($101$) and $s_6$ ($110$). The other automata within the clause take more reinforcement steps to converge their states diverging from this bias. This also explains the longer convergence time with lower number clauses. However, as the number of clauses is increased, the training convergence times decrease due with more redundancy and diversity between clauses.

Finally, we use an alternative means of fault mitigation by provisioning more states ($2n$) per automaton and study its impact on the reachability. For this, we repeat the stuck-at 1 fault injection in TA$_1$ register in bit position 0 for 4 different state sizes: from 6 to 12, each with a 4-clause configuration. Figure~\ref{fig:dep-states} shows the maximum training accuracy as well as their convergence times. As can be seen, the accuracy increases from 75\% to 100\% when the number of states is increased from 6 to 8, corresponding to a 1-bit increase in the automaton register size (Figure~\ref{fig:dep-states}(a)). High $2n$ allow each automaton to explore a larger state-space. Note that, with one clause unable to provide correct outcomes, the 6-state automaton converges faster than the 8-state automaton. However, as more state values are allowed in the automaton, the learning converges faster to the maximum accuracy of 100\% (Figure~\ref{fig:dep-states}(b)).
	\begin{landscape}
\begin{table*}[htb]
\small{
    \centering
    \caption{Analysis of Nash equilibria for 2-input XOR}
    \begin{tabular}{c||l|l|l|l||c}
         \#&$\mathbf{TA}_1$ -- $x_1$&$\mathbf{TA}_2$ -- $\lnot x_1$&$\mathbf{TA}_3$ -- $x_2$& $\mathbf{TA}_4$ -- $\lnot x_2$& \textbf{Clause}\\
         \hline
         \hline
         1.&Exclude ($-0.625$)&Exclude ($-0.625$)&Exclude ($-0.625$)&Exclude ($-0.625$)& $1$\\
         2.&Exclude ($-0.375$)&Exclude ($0.125$)&Exclude ($-0.1875$)&Include ($0.125$)&$\lnot x_2$\\
         3.&Exclude ($0.0625$)&Exclude ($-0.375$)&Include ($0.125$)&Exclude ($-0.1875$)&$x_2$\\
         4.&Exclude ($0.125$)&Exclude ($0.125$)&Include ($-0.125$)&Include ($-0.125$)&$x_2 \land \lnot x_2$\\
         5.&Exclude ($-0.1875$)&Include ($0.125$)&Exclude ($-0.375$)&Exclude ($0.0625$)&$\lnot x_1$\\
         6.&Exclude ($-0.125$)&Include ($-0.125$)&Exclude ($-0.125$)&Include ($-0.125$)&$\lnot x_1 \land \lnot x_2$\\
         \hline
         \bf{7.}&Exclude ($0.0625$)&Include ($0.125$)&Include ($0.125$)&Exclude ($0.0625$)&$\lnot x_1 \land x_2$\\
         \hline
         8.&Exclude ($0.125$)&Include ($-0.125$)&Include ($-0.125$) &Include ($-0.125$)&$\lnot x_1 \land x_2 \land \lnot x_2$\\
         9.&Include ($0.125$)&Exclude ($-0.1875$)&Exclude ($0.125$)&Exclude ($-0.375$)&$x_1$\\
         \hline
         \bf{10.}&Include ($0.125$)&Exclude ($0.0625$)&Exclude ($0.0625$)&Include ($0.125$)&$x_1 \land \lnot x_2$\\
         \hline
         11.&Include ($-0.125$)&Exclude ($-0.125$)&Include ($-0.125$)&Exclude ($-0.125$)&$x_1 \land x_2$\\
         12.&Include ($-0.125$)&Exclude ($0.125$)&Include ($-0.125$)&Include ($-0.125$)&$x_1 \land x_2 \land \lnot x_2$\\
         13.&Include ($-0.125$)&Include ($-0.125$)&Exclude ($0.125$)&Exclude ($0.125$)&$x_1 \land \lnot x_1$\\
         14.&Include ($-0.125$)&Include ($-0.125$)&Exclude ($0.125$)&Include ($-0.125$)&$x_1 \land \lnot x_1 \land \lnot x_2$\\
         15.&Include ($-0.125$)&Include ($-0.125$)&Include ($-0.125$)&Exclude ($0.125$)&$x_1 \land \lnot x_1 \land x_2$\\
         16.&Include ($-0.125$)&Include ($-0.125$)&Include ($-0.125$)&Include ($-0.125$)&$x_1 \land \lnot x_1 \land x_2 \land \lnot x_2$
    \end{tabular}
    \label{tab:game_theoretic_analysis}
    }
\end{table*}
\end{landscape}
\section{Game Theoretic Convergence Analysis}\label{sec:game_theoretic}

The stochasticity in \ac{TM} learning comes from (i) the random arrival of training samples, (ii) the random selection of clauses for updating, and (iii) the random generation of the rewards and penalties of Type I feedback. Here, we use a game-theoretic approach to analyze how Type I and Type II feedback guide the team of \ac{TA}s associated with a clause towards the optimal action configuration \cite{Granmo2018}. For this analysis we use a 2-input XOR, assuming uniformly distributed inputs, $P(x_1=1) = P(x_2=1) = 0.5$.

A game of \ac{TA} involves multiple automata and is played over several rounds~\cite{Narendra-book,Thathachar2004}. A round starts with each \ac{TA} deciding upon an action, which taken together govern rewarding of the individual \ac{TA}s. Responding to the rewards, the \ac{TA}s perform a random walk over the joint state space~\cite{Tung1996}. The interaction between the \ac{TA}s can thus be accurately represented by the payoff matrix of the game \cite{Narendra-book, Thathachar2004}.

A general analysis of the payoff matrix of the \ac{TM} can be found in \cite{Granmo2018}, while Table~\ref{tab:game_theoretic_analysis} contains the payoff matrix for 2-input XOR. The table shows 4 \ac{TA}s navigate the joint action space to produce a clause, through trajectories that always lead to the optimal action configuration. Each row  of the table specifies an action configuration. By assigning rewards a value of $1$ and penalties a value of $-1$, we can calculate the expected payoff of each \ac{TA}, per action configuration. The calculation is based on the Type I and Type II feedback tables (\cref{tab:feedback}), assuming uniformly distributed inputs and an $s$-value of $4$. Along with each action, in parentheses, the expected payoff of the action is listed. 

We have a Nash equilibrium if none of the \ac{TA}s can do better by unilaterally switching action \cite{Narendra-book,von2007theory}. For example, for the action configuration in row 1 of the table, $\mathrm{TA}_1$ has selected \textit{exclude}, which provides the expected payoff $-0.625$. The \ac{TA} would thus benefit by instead selecting \textit{include}, jumping to the action configuration of row 9. This would give an expected payoff of $0.125$ instead. Accordingly, the action configuration of row 1 is not a Nash equilibrium. 

By nature, a \ac{TA} pursues the action with the highest expected payoff (see Section~\ref{sec:reachability}), in this way seeking one of the Nash equilibria of the payoff matrix. However, if the probability of receiving a reward is smaller than that of receiving a penalty (the expected payoff is negative), the action is rejected~\cite{Narendra-book, Tsetlin1962}. Thus, the \ac{TA}s only accept a Nash equilibrium if all of the actions have positive expected payoff. From Table \ref{tab:game_theoretic_analysis}, this is the case only for the action configurations of row 7 and row 10. Both configurations produce a clause appropriate for the 2-input XOR problem. All of the other action configurations introduce prediction errors. However, these are transient configurations because they all contain actions with negative expected payoffs, thus repelling the \ac{TA}s to other configurations for convergence. Hence, the power of the scheme!


	\section{Summary and Conclusions} \label{sec:conclusions}
We presented the first insights into energy-frugality and explainability of learning automata based AI hardware design using hyperparameter search and reachability analysis. Our key findings are as follows.

\textit{Energy-Frugality}: For datasets with minor inter-class correlations, low learning sensitivity ($s$) and learning threshold ($T$) hyperparameter values provide robust learning in \ac{TM} with less number of clauses, thus providing energy-frugality. However, when inter-class correlations persist, increasing $T$ is essential for providing higher stochastic variations (\ie{} prodigality) during learning to avoid over-fitting and learning instability. For ensuring high-fidelity stochastic variations at low-cost, the precision of low-level randomization circuits needs to be suitably optimized.

\textit{Explainability and Dependability}: With a bounded state-space, \ac{TM} can start from random initial \ac{TA} states and yet reach a learnt state with incremental, discrete-event reinforcements. The LA algorithm guarantees convergence towards this learnt state as confirmed by our game theoretic analysis. With suitably chosen redundant clauses or automation state register bit-width and thereby more prodigality, faults can be fully masked and reachability property can be retained without requiring any additional fault mitigation strategy. Compared with clause redundancy approach, expanding the state register sizes provides more energy-frugality.

Our future work includes the development of a formal explainability analysis tool with comprehensive fault injection campaigns and energy optimization mechanisms.

\section*{Acknowledgments}
The authors would like to gratefully acknowledge the funding support from the UK Northern Accelerator (ref: NACCF 220), Lloyds Registers Foundation (ref: 5thICON-12) and Norwegian Research Council (ref: AIEverywhere project).

\bibliographystyle{abbrv} 
\bibliography{ieee-config,neural,tsetlin}

\begin{thebibliography}{10}

\bibitem{ansari2019improving}
M.~S. Ansari, V.~Mrazek, B.~F. Cockburn, L.~Sekanina, Z.~Vasicek, and J.~Han.
\newblock Improving the accuracy and hardware efficiency of neural networks
  using approximate multipliers.
\newblock {\em IEEE Transactions on Very Large Scale Integration (VLSI)
  Systems}, 28(2):317--328, 2019.

\bibitem{beninirtns17}
L.~Benini.
\newblock Plenty of room at the bottom: Micropower deep learning for cognitive
  cyberphysical systems.
\newblock {\em RTNS Keynote}, 2017.

\bibitem{bergstra2012random}
J.~Bergstra and Y.~Bengio.
\newblock Random search for hyper-parameter optimization.
\newblock {\em Journal of machine learning research}, 13(2), 2012.

\bibitem{bhatt2020explainable}
U.~Bhatt, A.~Xiang, S.~Sharma, A.~Weller, A.~Taly, Y.~Jia, J.~Ghosh, R.~Puri,
  J.~M. Moura, and P.~Eckersley.
\newblock Explainable machine learning in deployment.
\newblock In {\em Proceedings of the 2020 Conference on Fairness,
  Accountability, and Transparency}, pages 648--657, 2020.

\bibitem{conti2018xnor}
F.~Conti, P.~D. Schiavone, and L.~Benini.
\newblock Xnor neural engine: A hardware accelerator ip for 21.6-fj/op binary
  neural network inference.
\newblock {\em IEEE Transactions on Computer-Aided Design of Integrated
  Circuits and Systems}, 37(11):2940--2951, 2018.

\bibitem{doran2017does}
D.~Doran, S.~Schulz, and T.~R. Besold.
\newblock What does explainable ai really mean? a new conceptualization of
  perspectives.
\newblock {\em arXiv preprint arXiv:1710.00794}, 2017.

\bibitem{garofalo2020pulp}
A.~Garofalo, M.~Rusci, F.~Conti, D.~Rossi, and L.~Benini.
\newblock Pulp-nn: accelerating quantized neural networks on parallel
  ultra-low-power risc-v processors.
\newblock {\em Philosophical Transactions of the Royal Society A},
  378(2164):20190155, 2020.

\bibitem{gel1962some}
I.~M. Gel'fand and M.~L. Tsetlin.
\newblock Some methods of control for complex systems.
\newblock {\em Russian Mathematical Surveys}, 17(1):95, 1962.

\bibitem{rahimi2019tsetlin}
S.~R. {Gorji}, O.-C. {Granmo}, A.~{Phoulady}, and M.~{Goodwin}.
\newblock {A Tsetlin Machine with Multigranular Clauses}.
\newblock In {\em Lecture Notes in Computer Science: Proceedings of the
  Thirty-ninth International Conference on Innovative Techniques and
  Applications of Artificial Intelligence (SGAI-2019)}, volume 11927. Springer
  International Publishing, 2019.

\bibitem{Granmo2018}
O.-C. Granmo.
\newblock {The Tsetlin Machine - A Game Theoretic Bandit Driven Approach to
  Optimal Pattern Recognition with Propositional Logic}.
\newblock {\em arXiv e-prints}, Apr. 2018.

\bibitem{granmo2007learning}
O.-C. Granmo, B.~J. Oommen, S.~A. Myrer, and M.~G. Olsen.
\newblock Learning automata-based solutions to the nonlinear fractional
  knapsack problem with applications to optimal resource allocation.
\newblock {\em IEEE Transactions on Systems, Man, and Cybernetics, Part B
  (Cybernetics)}, 37(1):166--175, 2007.

\bibitem{gunning2017explainable}
D.~Gunning.
\newblock Explainable artificial intelligence (xai).
\newblock {\em Defense Advanced Research Projects Agency (DARPA), nd Web}, 2,
  2017.

\bibitem{lei2020arithmetic}
J.~Lei, A.~Wheeldon, R.~Shafik, A.~Yakovlev, and O.-C. Granmo.
\newblock From arithmetic to logic based ai: a comparative analysis of neural
  networks and tsetlin machine.
\newblock In {\em 2020 27th IEEE international conference on electronics,
  circuits and systems (ICECS)}, pages 1--4. IEEE, 2020.

\bibitem{lim2021spontaneous}
D.-H. Lim, S.~Wu, R.~Zhao, J.-H. Lee, H.~Jeong, and L.~Shi.
\newblock Spontaneous sparse learning for pcm-based memristor neural networks.
\newblock {\em Nature communications}, 12(1):1--14, 2021.

\bibitem{Maheshwari2023}
S.~Maheshwari, T.~Rahman, R.~ShafikSenior~member, A.~Yakovlev, A.~Rafiev,
  L.~Jiao, and O.-C. Granmo.
\newblock Redress: Generating compressed models for edge inference using
  tsetlin machines.
\newblock {\em IEEE Transactions on Pattern Analysis and Machine Intelligence},
  pages 1--16, 2023.

\bibitem{mathew2014energy}
J.~Mathew, R.~Shafik, and D.~K. Pradhan.
\newblock {\em Energy-efficient fault-tolerant systems}.
\newblock Springer USA, 2014.

\bibitem{mileiko2020neural}
S.~Mileiko, T.~Bunnam, F.~Xia, R.~Shafik, A.~Yakovlev, and S.~Das.
\newblock Neural network design for energy-autonomous artificial intelligence
  applications using temporal encoding.
\newblock {\em Philosophical Transactions of the Royal Society A},
  378(2164):20190166, 2020.

\bibitem{Narendra74}
K.~S. {Narendra} and M.~A.~L. {Thathachar}.
\newblock Learning automata - a survey.
\newblock {\em IEEE Transactions on Systems, Man, and Cybernetics},
  SMC-4(4):323--334, July 1974.

\bibitem{Narendra-book}
K.~S. Narendra and M.~A.~L. Thathachar.
\newblock {\em Learning Automata: An Introduction}.
\newblock Prentice-Hall, Inc., Upper Saddle River, NJ, USA, 1989.

\bibitem{Oommen1988}
B.~J. {Oommen} and J.~P.~R. {Christensen}.
\newblock $\epsilon$-optimal discretized linear reward-penalty learning
  automata.
\newblock {\em IEEE Transactions on Systems, Man, and Cybernetics},
  18(3):451--458, 1988.

\bibitem{qiqieh2018significance}
I.~Qiqieh, R.~Shafik, G.~Tarawneh, D.~Sokolov, S.~Das, and A.~Yakovlev.
\newblock Significance-driven logic compression for energy-efficient multiplier
  design.
\newblock {\em IEEE Journal on Emerging and Selected Topics in Circuits and
  Systems}, 8(3):417--430, 2018.

\bibitem{rahman2022data}
T.~Rahman, A.~Wheeldon, R.~Shafik, A.~Yakovlev, J.~Lei, O.-C. Granmo, and
  S.~Das.
\newblock Data booleanization for energy efficient on-chip learning using logic
  driven ai.
\newblock In {\em 2022 International Symposium on the Tsetlin Machine (ISTM)},
  pages 29--36. IEEE, 2022.

\bibitem{serb2018seamlessly}
A.~Serb, A.~Khiat, and T.~Prodromakis.
\newblock Seamlessly fused digital-analogue reconfigurable computing using
  memristors.
\newblock {\em Nature communications}, 9(1):1--7, 2018.

\bibitem{shafik2018real}
R.~Shafik, A.~Yakovlev, and S.~Das.
\newblock Real-power computing.
\newblock {\em IEEE Transactions on Computers}, 67(10):1445--1461, 2018.

\bibitem{shafik2009soft}
R.~A. Shafik, B.~M. Al-Hashimi, S.~Kundu, and A.~Ejlali.
\newblock Soft error-aware voltage scaling technique for power minimization in
  application-specific multiprocessor system-on-chip.
\newblock {\em Journal of Low Power Electronics}, 5(2):145--156, 2009.

\bibitem{shafik2008systemc}
R.~A. Shafik, P.~Rosinger, and B.~M. Al-Hashimi.
\newblock {SystemC}-based minimum intrusive fault injection technique with
  improved fault representation.
\newblock In {\em 14th IEEE Intl. On-Line Testing Symposium}, pages 99--104,
  2008.

\bibitem{Thathachar2004}
M.~A.~L. Thathachar and P.~S. Sastry.
\newblock {\em {Networks of Learning Automata: Techniques for Online Stochastic
  Optimization}}.
\newblock Kluwer Academic Publishers, 2004.

\bibitem{Tsetlin1962}
M.~L. Tsetlin.
\newblock {On the Behavior of Finite Automata in Random Media}.
\newblock {\em Automation and Remote Control}, 22:1210--1219, 1962.

\bibitem{Tung1996}
B.~Tung and L.~Kleinrock.
\newblock {Using Finite State Automata to Produce Self-Optimization and
  Self-Control}.
\newblock {\em IEEE Transactions on Parallel and Distributed Systems},
  7(4):47--61, 1996.

\bibitem{von2007theory}
J.~Von~Neumann and O.~Morgenstern.
\newblock {\em Theory of games and economic behavior (commemorative edition)}.
\newblock Princeton university press, 2007.

\bibitem{wang2019deep}
E.~Wang, J.~J. Davis, R.~Zhao, H.-C. Ng, X.~Niu, W.~Luk, P.~Y. Cheung, and
  G.~A. Constantinides.
\newblock Deep neural network approximation for custom hardware: Where we've
  been, where we're going.
\newblock {\em ACM Computing Surveys (CSUR)}, 52(2):40, 2019.

\bibitem{Wheeldon2020a}
A.~Wheeldon, R.~Shafik, T.~Rahman, J.~Lei, A.~Yakovlev, and O.-C. Granmo.
\newblock Learning automata based energy-efficient {AI} hardware design for
  {IoT} applications.
\newblock {\em Philosophical Trans. A of the Royal Society}, (in press), 2020.

\bibitem{yampolskiy2019unexplainability}
R.~V. Yampolskiy.
\newblock Unexplainability and incomprehensibility of artificial intelligence.
\newblock {\em arXiv preprint arXiv:1907.03869}, 2019.

\bibitem{zhang2020survey}
Y.~Zhang, P.~Ti{\v{n}}o, A.~Leonardis, and K.~Tang.
\newblock A survey on neural network interpretability.
\newblock {\em arXiv preprint arXiv:2012.14261}, 2020.

\end{thebibliography}

\end{document}